\pdfoutput=1
\documentclass[11pt]{article}

\usepackage{acl}
\usepackage{times}
\usepackage{latexsym}
\usepackage[T1]{fontenc}
\usepackage{graphicx}
\usepackage{amsfonts}
\usepackage{amsmath}
\usepackage[utf8]{inputenc}
\usepackage{microtype}
\usepackage{lipsum}
\usepackage{subcaption}
\usepackage{multirow}
\usepackage{algorithm2e}
\usepackage{pgfplots}
\usepackage{adjustbox}
\usepackage{url}
\usepackage{tikz}
\usepackage{xcolor}
\usepackage{comment}
\usetikzlibrary{patterns}
\usepackage{algorithmic}
\usepackage{comment}

\newcommand\tab[1][0.2cm]{\hspace*{#1}}

\definecolor{Original}{HTML}{c0bebd}
\definecolor{HighModel}{HTML}{33a055}
\definecolor{HighDictio}{HTML}{f2bc15}
\definecolor{LowModel}{HTML}{449bed}
\definecolor{LowDictio}{HTML}{e23334}

\usepackage[disable]{todonotes}

\title{Less is KEN: a Universal and Simple Non-Parametric Pruning Algorithm for Large Language Models}

\author{Michele Mastromattei \\
  University of Rome Tor Vergata\\
  Campus Bio-Medico University of Rome \\
  \texttt{michele.mastromattei@uniroma2.it} \\\And
  Fabio Massimo Zanzotto \\
  University of Rome Tor Vergata\\}

\begin{document}
\maketitle
\begin{abstract}
Neural network pruning has become increasingly crucial due to the complexity of these models and their widespread use in various fields. Existing pruning algorithms often suffer from limitations such as architecture specificity, excessive complexity and reliance on demanding calculations, rendering them impractical for real-world applications.
This paper introduces KEN: a straightforward, universal and unstructured pruning algorithm based on Kernel Density Estimation (KDE). KEN\footnote{Code available at \url{https://github.com/itsmattei/KEN}} aims to construct optimized transformers by selectively preserving the most significant parameters while restoring others to their pre-training state. This strategy preserves model performance while enabling storage of only the optimized subnetwork, leading to substantial memory savings.
Extensive evaluations across seven different LLMs demonstrate that KEN achieves equal or better performance than their original unpruned versions, with a minimum parameter reduction of 25\%. Furthermore, in-depth comparisons with established pruning and PEFT algorithms confirm KEN effectiveness. We further introduce KEN$_{viz}$, an explainable tool that visualizes the optimized model composition achieved by KEN from different points of view.

\end{abstract}

\section{Introduction}

Large Language Models (LLMs) have become the best and simplest solution for achieving state-of-the-art results in many natural language processing (NLP) applications. However, the increasing use of neural networks (NNs) and transformers \cite{vaswani2017attention} has resulted in a rise in computational cost due to the complexity of arithmetic calculations, larger matrices and the addition of more layers. Consequently, the weight and structure of these models become more complex, requiring high demands in computation and memory.

One of the best approaches to address the overwhelming size of LLMs is to reduce their resources through \textit{pruning algorithms}. These algorithms can eliminate parameters or entire components in a NN, making it lighter without compromising its original performance. Pruning algorithms emerged in parallel with the earliest use of NNs \cite{mozer1989using, janowsky1989pruning, lecun1989optimal}, but they have gained significant importance in the last decade due to the widespread use of these networks in various fields. There are many pruning algorithms in literature \cite{blalock2020state}, each with a unique approach or adapted old algorithms for these new architectures \cite{benbaki2023fast}.
However, the complexity of neural networks can pose a challenge when creating pruning algorithms, as these may require new complex theorems to make the models lightweight \cite{dong2017learning, malach2020proving}. 
Additionally, existing pruning algorithms often exhibit shortcomings in their completeness \cite{blalock2020state} and fail to consider a critical aspect: the efficient storage of the pruned result.
Some algorithms compress models at runtime but lack mechanisms to preserve the reduced NN for future use. Consequently, most algorithms prioritize the speed of reduction and execution, neglecting this critical final stage essential in resource-limited environments \cite{yang2017designing, sze2017efficient}.

\begin{figure}[h!]
    \centering
    \includegraphics[width=\linewidth]{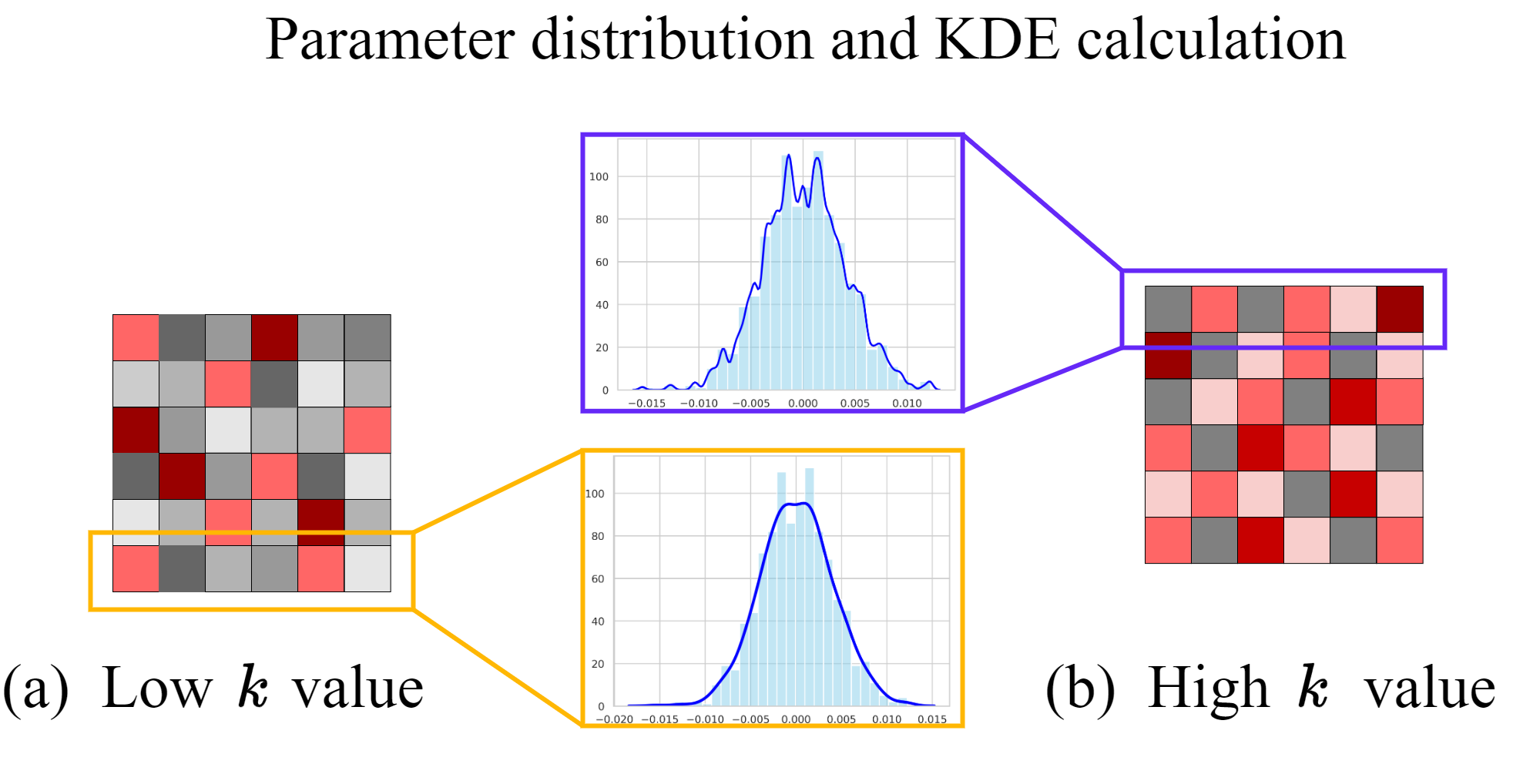}
    \caption{How $k$ value influences the KDE calculation, driving the parameter selection}
    \label{fig: different distribution}
\end{figure}

This paper presents KEN (\textbf{K}ernel density \textbf{E}stimator for \textbf{N}eural network compression): a universal, simple, magnitude-based transformer pruning algorithm that leverages Kernel Density Estimation (KDE) for parameter pruning. Unlike other pruning methods that rely on minimizing loss functions or exhaustive parameter search, KEN - inspired by the \textit{winning ticket} pruning hypothesis \cite{frankle2018lottery} - identifies and retains the most influential parameters using KDEs while resetting the others to their original pre-trained values. This innovative approach streamlines the optimization process by leveraging the natural distribution of model parameters, eliminating any architecture-specific considerations.
KEN effectively reduces the size of transformer models by a minimum of 25\% without compromising performance. The pruned models consist only of a subnetwork of trained parameters, which can be seamlessly downloaded and injected into its pre-trained version as needed. This feature enables dynamic model reconfiguration and saves significant memory space that would otherwise be required to store the fully trained model.
Comparative evaluations demonstrate KEN exceptional capabilities, surpassing existing transformer pruning and PEFT algorithms. Finally, we introduce KEN$_{viz}$: an explainable tool that graphically depicts the optimized model from various perspectives. KEN$_{viz}$ highlights the KEN-selected parameters, their layer-wise differences and neighbor counts for each matrix that made up the analyzed model.
Using KEN, we employed a non-parametric method widely used in statistics, to create an efficient and intuitive pruning algorithm. Our approach achieved excellent results in terms of efficiency and performance, making it a practical alternative to other more complex pruning algorithms.

\section{Background}
Compression algorithms can be summarized in three areas of research: weight pruning \cite{han2015learning, zhu2017prune}, quantization \cite{gong2014compressing, zhu2016trained} and knowledge distillation \cite{ba2014deep, kim2016sequence}. These techniques aim to make models lighter, but each of them takes a different approach. Weight pruning removes model parameters according to the chosen algorithm and strategy, while quantization reduces the number of bits necessary to represent each parameter. Knowledge distillation, instead, tries to minimize the learned large knowledge of a model into a smaller one without affecting its \textit{validation}.

Focusing on pruning algorithms, there are different approaches depending on the strategy and algorithm adopted. Pruning algorithms can be classified as either \textit{structured} or \textit{unstructured}, based on the approach applied and \textit{magnitude-based} or \textit{impact-based}, according to the algorithm used. Structured pruning \cite{huang2018condensenet, wang2019structured, gordon2020compressing} removes weights in groups, such as entire neurons, filters or layers, while unstructured pruning \cite{han2015learning, frankle2018lottery, lagunas2021block, benbaki2023fast} does not consider any relationship between parameters and selects weights to prune based on their impact or magnitude. Magnitude-based algorithms \cite{hanson1988comparing, mozer1989using, gordon2020compressing} analyze the absolute value of each parameter to determine its importance. In contrast, impact-based algorithms \cite{lecun1989optimal, hassibi1992second, singh2020woodfisher} work on the loss function and its variation caused by removing a parameter.
The \textit{winning ticket} hypothesis \cite{frankle2018lottery}, is a recent advancement in pruning techniques. A \textit{winning ticket} is a subnetwork within a trained model that - when trained in isolation - can achieve performance comparable to the original model even after significant pruning. To identify the winning ticket, a pruning criterion is applied to zero-mask weights and the remaining network is retrained. This process can be repeated multiple times or in a one-shot manner.

\begin{figure*}[thb!]
    \centering
    \includegraphics[width=\linewidth]{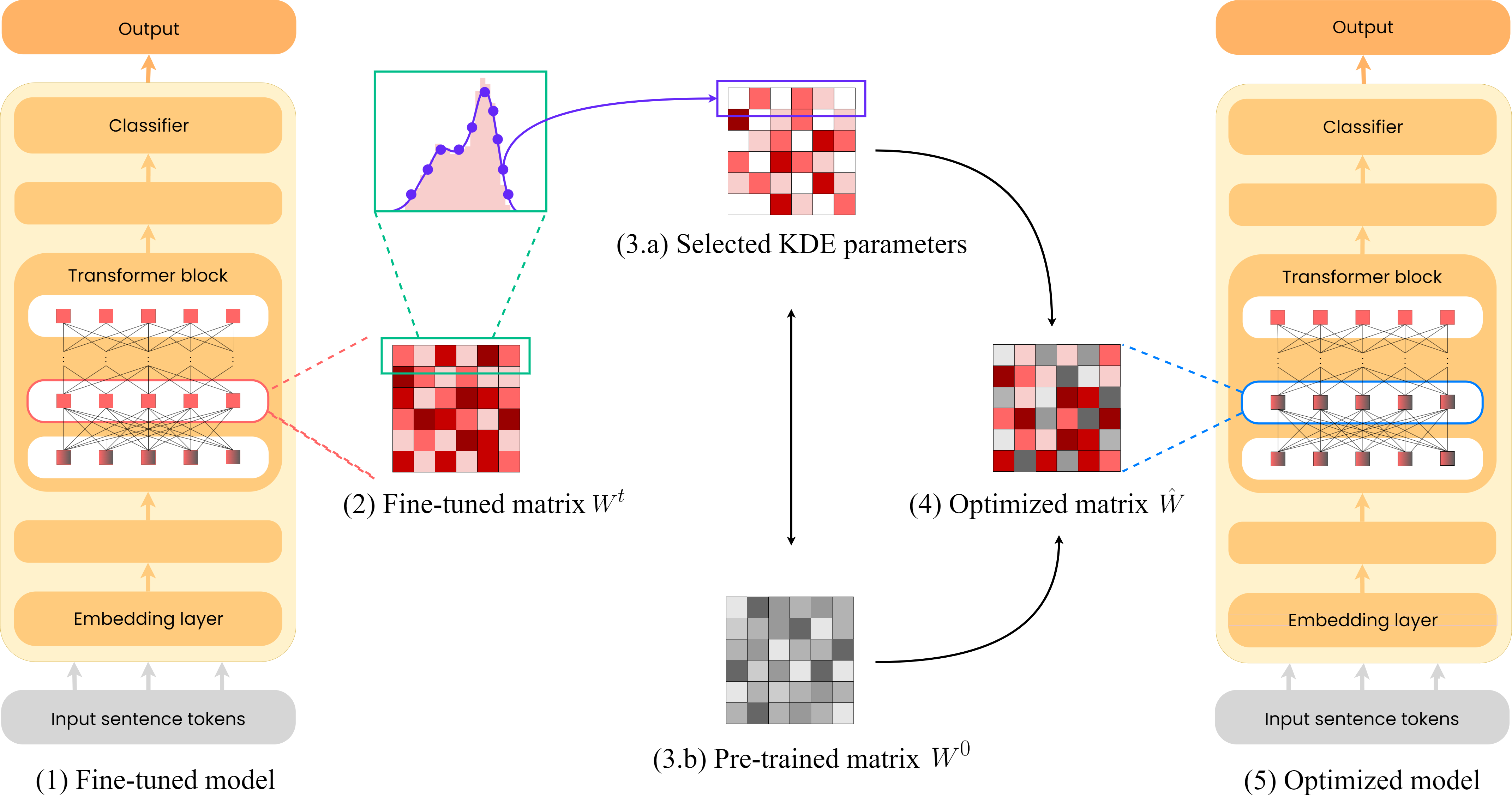}
    \caption{KEN workpath: From a fine-tuned model (1), for each of its fine-tuned matrices (2), the row distribution and the respective KDE (Kernel Density Estimator) are calculated. All values within the KDE are selected (3.a), while the remainder are restored to their pre-tuned value (3.b). The resulting optimized matrix (4) is then fed back into the model (5)}
    \label{fig:workflow}

\end{figure*}

\section{Related Work} \label{sec: related_work}
In this section, we present three algorithms that are relevant benchmarks for our proposed algorithm, KEN. These algorithms have some similarities with it: the first two, called FLOP and BMP, are pruning algorithms designed to reduce the size of transformer models by employing algebraic or geometric techniques. The third, LoRA is the SoTA parameter-efficient algorithm for LLMs.

\paragraph{Factorized Low-rank Pruning} (FLOP: \citealp{wang2019structured}) is a magnitude-based pruning algorithm that employs matrix factorization to reduce the size of matrices in transformer models. This approach involves decomposing each matrix into rank components, which are then multiplied together to form the original matrix. For attention layers, FLOP decomposes each matrix into smaller rank-1 components based on the magnitudes of the matrix entries. Instead, for embedding layers, FLOP adaptively prunes dimensions based on word clusters. This means that FLOP only prunes dimensions that are not frequently used \cite{joulin2017efficient, bastings2019interpretable}, which helps to reduce the model size without sacrificing performance.

\paragraph{Block Movement Pruning} (BMP: \citealp{lagunas2021block}) introduces an extension to the movement pruning technique used in transformers \cite{sanh2020movement}. This approach reduces the size of each matrix in a transformer model by dividing it into fixed-sized blocks. Regularization is then applied, and the NN is trained through distillation to match the performance of a teacher model. Our focus is on two pruning methods: Hybrid and HybridNT. The key difference between these two approaches is that HybridNT does not involve the use of a teacher model during training (No Teacher).

\paragraph{Low-Rank Adaptation of Large Language Models} (LoRA: \citealp{hu2021lora}) is a novel fine-tuning method that leverages low-rank decomposition to reduce the parameter size of LLMs while preserving their performance. This approach involves decomposing LLMs weight matrices into low-rank components, which are then fine-tuned along with the original weights. LoRA enables efficient parameter adaptation to specific tasks without compromising the model generalization capabilities.

\section{KEN pruning algorithm} \label{subsec: KEN algorithm}
KEN (\textbf{K}ernel density \textbf{E}stimator for \textbf{N}eural network compression) pruning algorithm is designed to identify and extract the most essential subnetwork from transformer models following the main idea of the \textit{winning ticket} hypothesis \cite{frankle2018lottery}. Our algorithm effectively prunes the network by employing Kernel Density Estimators (KDEs), retaining only the essential parameters and resetting the rest to their pre-trained values. The optimized subnetwork can be stored independently and seamlessly integrated into its pre-trained configuration for downstream applications.

KEN through KDEs, generalizes the point distribution of each transformer matrix, capturing the \textit{smoothed} version of the original fine-tuned model. To prevent the complete deconstruction of the initial matrix composition, KEN applies KDEs to individual rows. The KDE calculation requires a $k$ value, which defines the number of points employed in the distribution calculation which directly influences the number of retained fine-tuned parameters (Fig. \ref{fig: different distribution}). Thus, a lower $k$ value indicates a closer resemblance to the pre-trained model while a higher $k$ value reflects a closer alignment with its fine-tuned version.

KEN algorithm operates in three primary steps:

\paragraph{Step 1: Parameter Extraction and KDE Calculation} Given a pre-trained matrix $W^{0}$ of a layer $l$: 
\begin{equation*}
    W^{0} =  \{w_{1,1}^0, ..., w_{n,m}^0\} \quad | \quad W^{0}\in \mathbb{R}^{n \times m}
\end{equation*}
and its corresponding fine-tuned counterpart $W^{t}$:
  \begin{equation*}
    W^{t} = \{w_{1,1}^{t}, ..., w_{n,m}^{t}\} \quad | \quad W^{t} \in \mathbb{R}^{n \times m}
\end{equation*}
for each row $r_i^t$ of the fine-tuned matrix $W^t$:
\begin{equation*}
    r_{i}^t = \{w_{i,1}^t, ... , w_{i,m}^t\} \quad \forall i \in [1,n]
\end{equation*}
KEN calculates the KDE distribution of the row $r_i^t$ using a bandwidth parameter $h$ determined following Scott’s rule of thumb \cite{scott2015multivariate}.
\begin{equation*}
    h = 1.06 \cdot \hat{\sigma}\cdot n^{-\frac{1}{5}}
\end{equation*}
where $\hat{\sigma}$ is the standard deviation of $r_i^t$.
\paragraph{Step 2: Parameter Retention and Pre-trained Value Reset} 
The $k$ points that best fit the $r_i^t$ row distribution are identified using the KDE likelihood, while the others are reset to their pre-trained values. This process results in an optimized row $\hat{r}_i$:

\begin{equation*}
    \hat{r}_{i} = \{\hat{w}_{i,1}, ... , \hat{w}_{i,m}\} \quad \forall i \in [1,n]
\end{equation*}
computed using the following binary function:
\begin{equation}
f(\hat{w}_{i,j}) = 
    \begin{cases}
    w_{i,j}^t \tab $if$ \quad w_{i,j}^t \in \tab $KDE \tab   likelihood$\\
    w_{i,j}^0 \quad $otherwise$
\end{cases}\
\end{equation}
\paragraph{Step 3: Matrix Replacement and Optimized Fine-tuned Model} After applying the previous step on each row, the optimized matrix $\hat{W}$:
\begin{equation*}
    \hat{W} = \{ \hat{w}_{1,1}, ..., \hat{w}_{n,m} \} \quad | \quad \hat{W} \in \mathbb{R}^{n \times m}
\end{equation*}
will replace the original fine-tuned matrix $W^t$ within the model.

KEN operates iteratively, replacing the $W^t$ matrix with $\hat{W}$ during each iteration. Therefore, after the $t-th$ iteration, the model will have $t-optimized$ matrices, effectively replacing the fine-tuned matrices without creating any additional versions of the model. This versatility allows KEN to prune the entire model or specific layer ranges.

\RestyleAlgo{ruled}
\begin{algorithm}
\caption{Generate the optimized $\hat{W}$ matrix using KEN}\label{alg: Define sparse matrix}
\KwData{$W^{0} = \{w_{1,1}^0, ..., w_{n,m}^0\}$, $W^{t} = \{w_{1,1}^{t}, ..., w_{n,m}^{t}\}, k$}
\KwResult{$\hat{W}$}
$\hat{W}[n,m] \gets 0 $

\For{i = 1 to n}{
    $\texttt{best\_points} \gets KDE(r_{i}^{t}, k)$
    
    \For{j = 1 to m}{
        $\hat{r}_{i}^{t} \gets []$
        
        \eIf{$r_{i}^{t}[j]$ in \texttt{best\_points}}{
        $\hat{r}_{i}^{t}[j] \gets r_{i}^{t}[j]$
        }{
        $\hat{r}_{i}^{t}[j]$$ \gets r_{i}^{0}[j]$
        }
    }
    $\hat{W}[i] \gets \hat{r}_{i}^{t} $
}
\Return $\hat{W}$
\end{algorithm}

\begin{figure*}[htb!]
     \centering
     \begin{subfigure}[b]{0.2\textwidth}
         \centering
         \includegraphics[width=\textwidth]{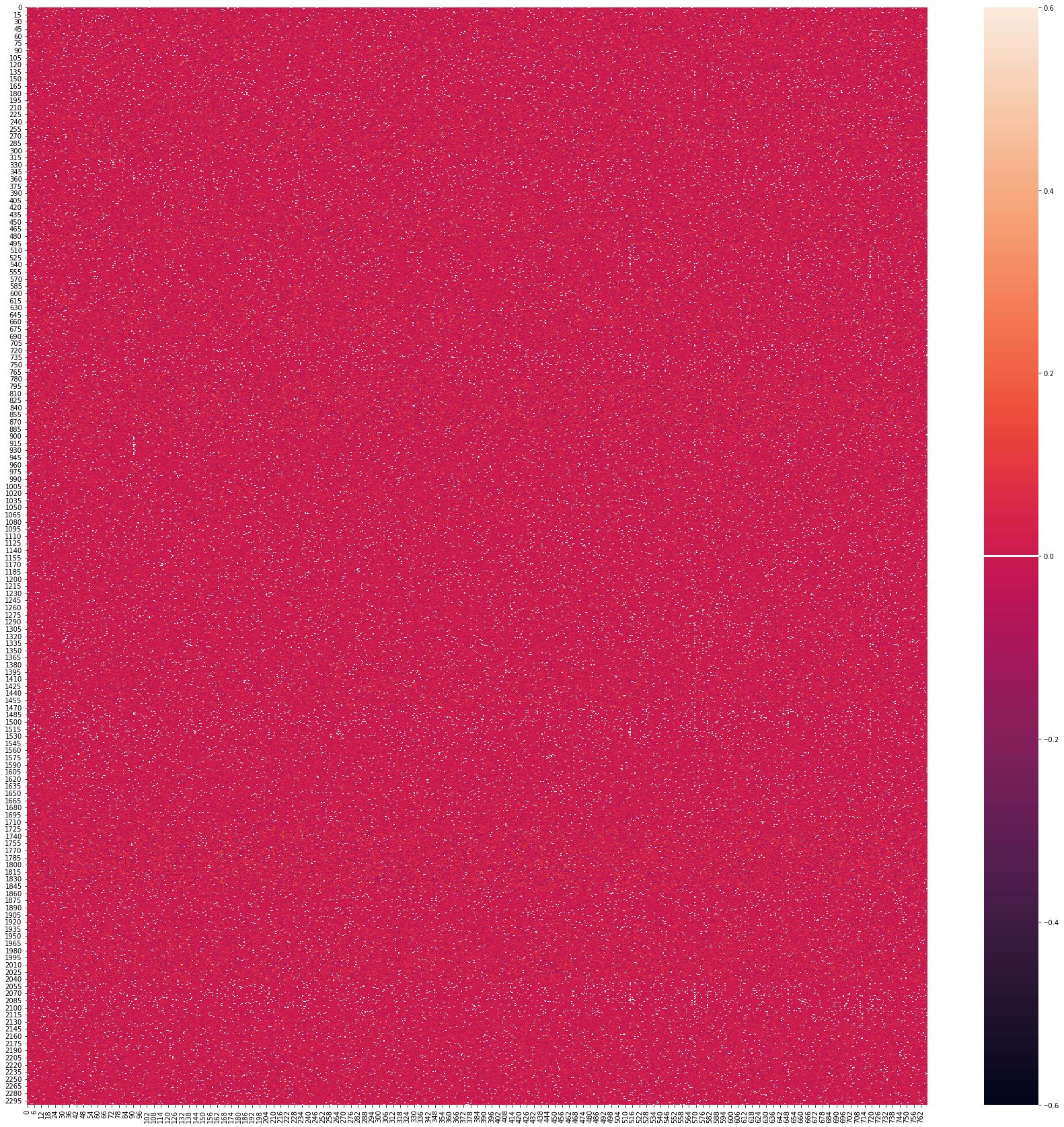}
         \caption{Fine-tuned matrix}
         \label{fig:Fine-tuned matrix}
     \end{subfigure}
     \hfill
     \begin{subfigure}[b]{0.2\textwidth}
         \centering
         \includegraphics[width=\textwidth]{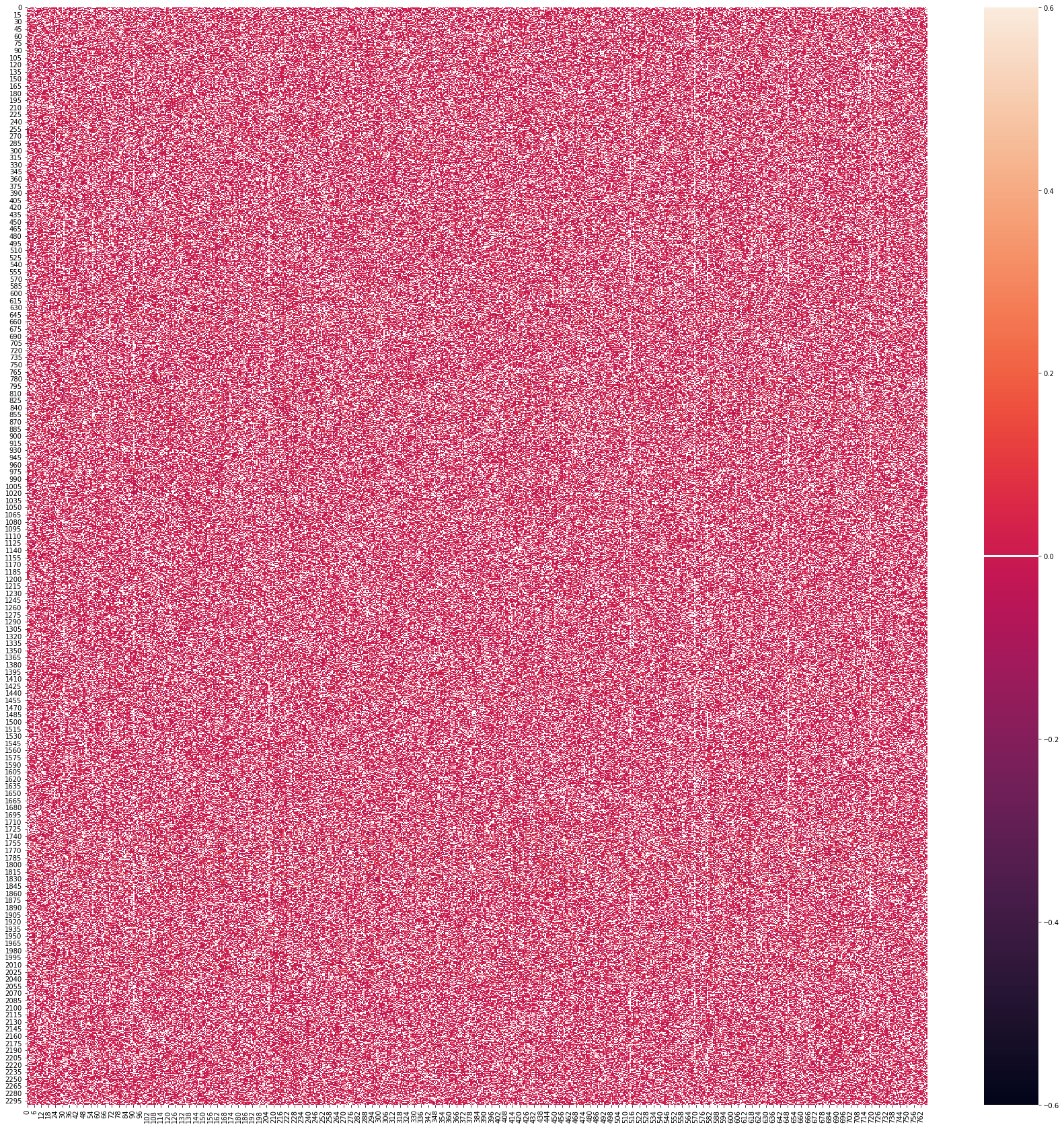}
         \caption{Reset params 34.90\%}
         \label{fig:Sparse matrix 500}
     \end{subfigure}
     \hfill
     \begin{subfigure}[b]{0.2\textwidth}
         \centering
         \includegraphics[width=\textwidth]{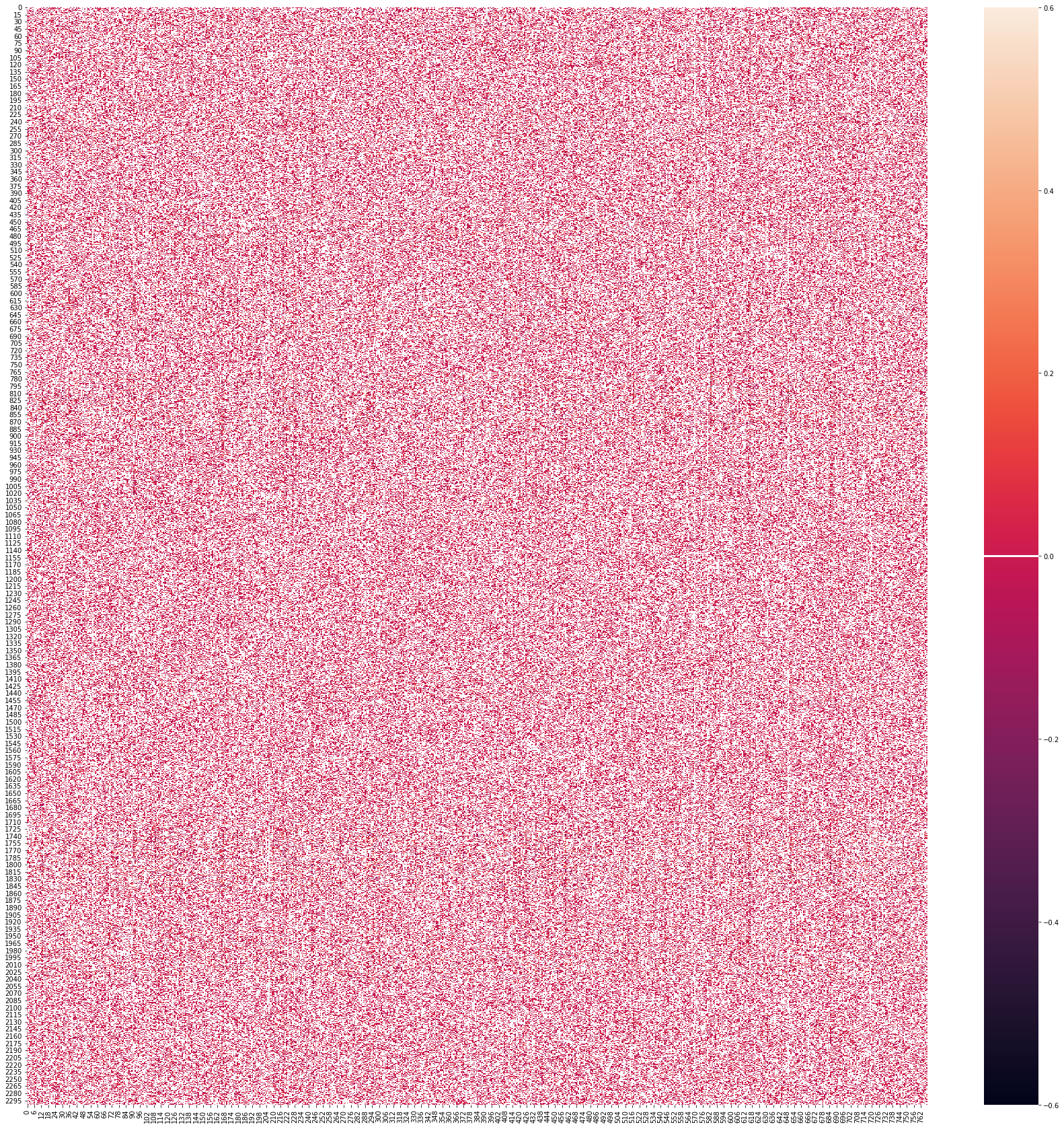}
         \caption{Reset params: 60.94\%}
         \label{fig:Sparse Matrix 300}
     \end{subfigure}
     \hfill
     \begin{subfigure}[b]{0.2\textwidth}
         \centering
         \includegraphics[width=\textwidth]{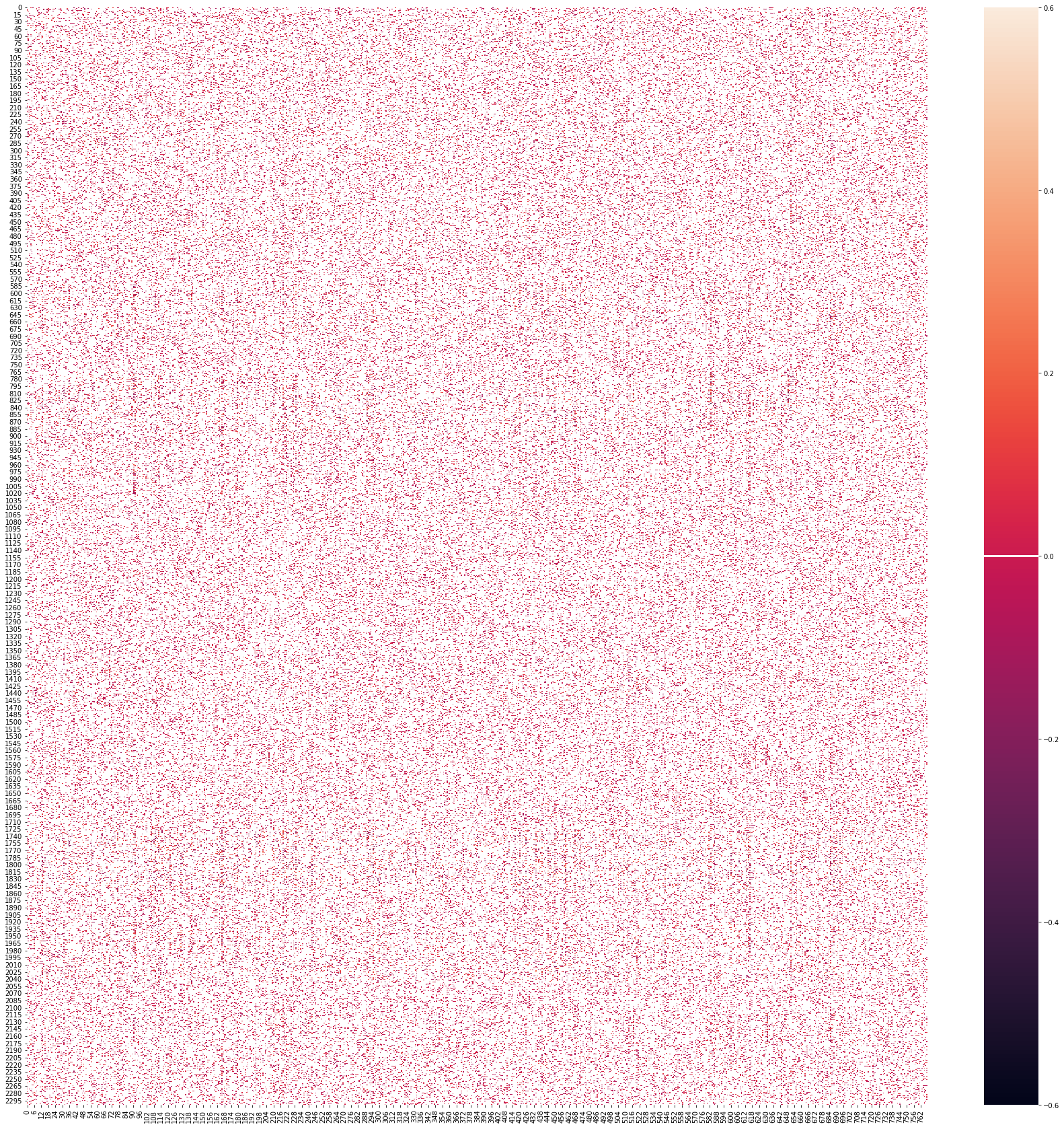}
         \caption{Reset params: 86.98\%}
         \label{fig:Sparse Matrix 100}
     \end{subfigure}
    
    \caption{Comparing the impact of KEN parameter selection on the same fine-tuned matrix (a). Matrix (a) represents the \texttt{in\_proj} matrix at layer 0 of a DeBERTa model trained on the \texttt{AG\_NEWS} dataset. No selected parameters are blank}
    \label{fig: Sparse carosel}
\end{figure*}

Algorithm \ref{alg: Define sparse matrix} provides a more formal explanation of the three steps described for generating the optimized matrix $\hat{W}$. Additionally, the graphical representation in Fig. \ref{fig:workflow} offers a clear and comprehensive visualization of all KEN steps, while Fig. \ref{fig: Sparse carosel} displays different $\hat{W}$ matrices obtained using various $k$ values.

\section{Experiments}
To validate our algorithm, we conducted a series of extensive case studies. Sec. \ref{subsec: Experimental set-up} describes the experimental setup, including the models employed and the $k$ values tested. Additionally, Sec. \ref{subsec: Model compression} focused on investigating the feasibility of saving and loading compressed data.

\subsection{Experimental set-up} \label{subsec: Experimental set-up}
To evaluate KEN pruning algorithm performance across different architectures and datasets, we conducted a thorough series of experiments using seven distinct transformer models. To maintain consistent evaluation conditions, we uniformly divided each dataset into training, validation and test sets. These divisions remained consistent throughout our experiments and across models. All datasets were imported from Huggingface\footnote{\url{https://huggingface.co/datasets}}. To achieve optimal performance, we fine-tuned each model before applying KEN algorithm per each dataset, adjusting the number of epochs until the fine-tuned model achieved the best F1-weighted score. Despite what the literature suggests, we used the F1 measure instead of classical accuracy as a comparison metric - if not explicitly used by the comparison benchmarks - because it delivers more reliable predictions, particularly on strongly unbalanced datasets. 

To fully assess KEN capabilities, we gradually increased the $k$ value required by the algorithm, starting from a low $k$ value and incrementally increasing it until its fine-tuned version was reached. This incremental approach allowed us to identify the critical \textit{threshold value} whereby the compressed model obtained results similar to its fine-tuned version or when the compression value $k$ leads to a catastrophic decline of performances, as reported in Apx. \ref{Appendix: random KEN}.

To provide a comprehensive analysis of KEN, we selected different transformer models with unique architecture, attention mechanisms, training approaches or different versions of the same model. Tab.\ref{tab: Table Model} compares the architectures of the models examined, emphasizing the number of layers and the number of parameters of each.

\begin{table}[h!]
    \centering
    \begin{adjustbox}{width=\linewidth}
    \begin{tabular}{{llcc}}
    \hline
        Model & & \# Layers & \# params \\ \hline
        BLOOM$_{1B7}$ & \cite{workshop2022bloom} & 24 & 1.72 B \\
        BLOOM$_{560k}$ &\cite{workshop2022bloom} & 24 & 560 M \\
        DeBERTa &\cite{he2020deberta} & 12        &  138 M\\
        Bert &\cite{devlin2018bert} &  12        &  109 M\\
        Ernie &\cite{sun2020ernie} & 12        &  109 M \\
        DistilBERT &\cite{sanh2019distilbert} & 6         &  66 M \\
        Electra &\cite{clark2020electra} & 12        &  33 M \\ \hline
    \end{tabular}
    \end{adjustbox}
    \caption{Properties of the analyzed models}
    \label{tab: Table Model}
\end{table}

\begin{table*}[htb!]
\centering
\begin{adjustbox}{width=\textwidth}
\begin{tabular}{lccccccc}
\hline
Model & Trainable params &  Reset params (\%) &  \texttt{AG-NEWS} &  \texttt{EMO}  &  \texttt{IMDB}  & \texttt{YELP\_POLARITY} &  \texttt{glue-sst2} \\ \hline

\multirow{3}{*}{BLOOM$_{1B7}$}     & 442M & 74.31 & 87.5 ($\pm 0.1$)   & \textbf{88.0} ($\pm 0.1$) &  76.6 ($\pm 0.1$) & \textbf{96.1 ($\pm 0.1$)} &   \textbf{80.4 ($\pm 0.1$)}     \\
                            & 531M & 69.17 & 92.2 ($\pm 0.1$)   & \textbf{90.6} ($\pm 0.1$) &  84.2 ($\pm 0.1$) & \textbf{96.3 ($\pm 0.1$)} &   \textbf{90.9 ($\pm 0.1$)}  \\
                            & 664M & 61.46 & \textbf{93.1 ($\pm 0.1$)}   & \textbf{90.1} ($\pm 0.1$) &  \textbf{87.6 ($\pm 0.1$)} &  \textbf{96.5 ($\pm 0.1$)} &    \textbf{92.9 ($\pm 0.1$)}  \\ \hline

\multirow{3}{*}{BLOOM$_{560k}$}     & 411M &26.34 & 91.3 ($\pm 0.1$)   & \textbf{81.4 ($\pm 0.1$)} &         \textbf{82.7 ($\pm 0.1$)} &     \textbf{95.2 ($\pm 0.1$)} &                  \textbf{92.4 ($\pm 0.1$)}         \\
                            & 420M & 24.80 & 91.8 ($\pm 0.1$)   & \textbf{83.0 ($\pm 0.1$)} &       \textbf{84.3 ($\pm 0.1$)} &     \textbf{95.3 ($\pm 0.1$)} &                 \textbf{92.1 ($\pm 0.1$)}    \\
                            & 429M &23.26 & \textbf{92.1 ($\pm 0.1$)}   & \textbf{84.0 ($\pm 0.1$)}&  \textbf{85.8 ($\pm 0.1$)} &     \textbf{95.3 ($\pm 0.1$)} &                 \textbf{92.3 ($\pm 0.1$)}          \\ \hline

\multirow{3}{*}{DeBERTa}    & 92M & 33.86 & 92.2 ($\pm 0.1$)           & \textbf{87.9 ($\pm 1.2$)}     & 82.5 ($\pm 5.1$)  & 95.9 ($\pm 0.4$)& 94.6 ($\pm 0.2$)\\
                            & 99M &  28.35 &\textbf{92.7 ($\pm 0.1$)}  & \textbf{87.3 ($\pm 1.0$)}     &  88.3 ($\pm 1.1$) & \textbf{96.1 ($\pm 0.2$)}& \textbf{94.9 ($\pm 0.1$)} \\
                            & 107M & 22.84 &\textbf{92.9 ($\pm 0.1$)}  & \textbf{87.1 ($\pm 1.2$)}     & \textbf{89.8 ($\pm 0.1$)}& \textbf{96.2 ($\pm 0.1$)}       &   \textbf{94.8 ($\pm 0.1$)}        \\ \hline

\multirow{3}{*}{Bert}       &  69M & 37.05 & \textbf{93.4 ($\pm 0.1$)}   & 84.2 ($\pm 1.1$)              & 86.8 ($\pm 0.1$) &  \textbf{95.0 ($\pm 0.4$)} & \textbf{93.7 ($\pm 0.5$) }         \\
                            & 75M & 31.80 &\textbf{93.7 ($\pm 0.2$)}   & \textbf{87.4 ($\pm 0.7$)}     & \textbf{87.3 ($\pm 0.1$)} & \textbf{95.0 ($\pm 0.5$)} & \textbf{93.7 ($\pm 0.4$)}    \\
                            & 80M & 26.55 &\textbf{93.6 ($\pm 0.1$)}   & \textbf{87.9 ($\pm 0.3$)}     & \textbf{87.6 ($\pm 0.1$)} & \textbf{95.1 ($\pm 0.4$)} & \textbf{93.8 ($\pm 0.4$)}          \\ \hline
                                                              
\multirow{3}{*}{Ernie}     & 69M & 37.05 &\textbf{93.3 ($\pm 0.4$)} & \textbf{89.1 ($\pm 0.6$)} & \textbf{89.4 ($\pm 0.2$)} & \textbf{95.8 ($\pm 0.1$)} &  \textbf{94.1 ($\pm 0.2$)}\\
                            & 75M & 31.80 &\textbf{93.3 ($\pm 0.3$)} & \textbf{88.7 ($\pm 1.2$)} & \textbf{89.2 ($\pm 0.2$)} &  \textbf{95.8 ($\pm 0.3$)} &  \textbf{93.8 ($\pm 0.2$)}\\
                            & 80M & 26.55 &\textbf{93.8 ($\pm 0.2$)} & \textbf{88.1 ($\pm 0.8$)} & \textbf{89.6 ($\pm 0.3$)} & \textbf{95.9 ($\pm 0.2$)} & \textbf{93.4 ($\pm 0.2$)} \\ \hline

\multirow{3}{*}{DistilBERT} & 44M & 34.39 & 92.3 ($\pm 0.6$)            & \textbf{88.1 ($\pm 1.4$)}     & 83.2 ($\pm 1.1$) & 94.6 ($\pm 0.1$) & \textbf{91.9 ($\pm 0.2$)} \\
                            & 47M & 28.92 & \textbf{93.1 ($\pm 0.2$)}   & \textbf{88.8 ($\pm 0.6$)}     & 84.4 ($\pm 0.5$) & 94.7 ($\pm 0.1$)& \textbf{91.9 ($\pm 0.1$)}\\
                            & 51M & 23.45 & \textbf{93.3 ($\pm 0.2$)}   & \textbf{88.2 ($\pm 0.3$)}     & \textbf{84.6 ($\pm 0.9$)} & \textbf{94.9 ($\pm 0.1$)}& \textbf{92.0 ($\pm 0.1$)} \\ \hline
                            
\multirow{3}{*}{Electra}    & 8.9M & 75.56 & 84.1 ($\pm 2.4$) &  84.3 ($\pm 0.4$) &  78.9 ($\pm 0.5$) & 88.5 ($\pm 0.9$)& 79.9 ($\pm 0.7$) \\
                            & 12M & 64.75 & 89.7 ($\pm 0.3$) & \textbf{86.0 ($\pm 0.3$)} &  82.0 ($\pm 0.5$) & 92.1 ($\pm 0.8$)& 85.0 ($\pm 0.2$)\\
                            & 14M & 55.94 & \textbf{91.3 ($\pm 0.2$)}  & \textbf{85.6 ($\pm 0.3$)} &  84.3 ($\pm 0.1$) & \textbf{93.7 ($\pm 0.4$)} &90.1 ($\pm 0.1$) \\ \hline
\end{tabular}
\end{adjustbox}
\caption{Results on various datasets obtained using different trainable parameters. Bold results indicate a similar or better F1-weighted value compared to the original (\textit{unpruned}) model. The reset params column indicates the percentage of the restored pre-trained params in the model. Other results are shown in Apx.\ref{Apx: Additional_results}}
\label{tab:Results}
\end{table*}

\subsection{Model compression} \label{subsec: Model compression}
Transformer models, like many neural networks, often have large file sizes. A fine-tuned transformer can range from 500 MB to 2GB or more. However, the KEN algorithm reduces this size by selecting and retaining a subset of $k$ parameters while restoring the rest to their pre-trained values. This process creates a more concentrated model that only includes the essential $k$ values for each matrix, resulting in significant weight reduction. To quantify the effectiveness of KEN, we save the compressed model generated during this phase and compare it to its original, unpruned version. To ensure a fair comparison, we use the same technique to save both the compressed and original fine-tuned models. However, KEN requires a support file, such as a dictionary, to load the $k$ parameters saved into their appropriate positions during the loading process. This is because during loading, the $k$ fine-tuning values must be loaded into a pre-trained model and the support file provides the necessary mapping to ensure proper placement. Sec. \ref{subsec: compression results} provides a comprehensive overview of the compression results obtained during this analysis.

\section{Results and Discussion} \label{sec: results}
In this section, we present the results obtained for each KEN main goals. Sec. \ref{subsec: result and comparison} discusses the effectiveness of KEN-pruned models compared to their unpruned counterparts, pruning benchmarks and the state-of-the-art PEFT algorithm. Sec. \ref{subsec: compression results} focuses on the process of saving and loading the subnetwork extracted by KEN, comparing the reduced file sizes achieved by it with those of the original models. Finally, Sec. \ref{subsec:KEN_viz} shows KEN$_{viz}$, illustrating its applications.

\subsection{Experiment results} \label{subsec: result and comparison}
To evaluate the efficacy of KEN, we conducted a series of experiments across different classification and sentiment analysis datasets. For each dataset, we implemented KEN multiple times, employing varied $k$ values and calculating the mean and standard deviation of the resulting F1-weighted scores. The complete dataset list can be found in Apx. \ref{Apx: Additional_results}. As evidenced in Tab. \ref{tab:Results}, KEN successfully compressed all analyzed models without sacrificing their original, unpruned performance. We observed a remarkable reduction in overall model parameter count, ranging from a minimum of 25\% to a substantial $\approx$ 70\% for certain models. Intriguingly, the models with both the highest and lowest parameter counts exhibited the most significant parameter reduction. Additionally, for each model under examination, we observed no substantial difference in performance as the percentage of reset parameters increased, maintaining a remarkable resemblance to the unpruned model performance. This observation underscores KEN exceptional generalization capability, balancing performance and compression even at middle-high compression rates.

We compared KEN to other pruning algorithms specifically designed for transformer models, including FLOP, Hybrid and HybridNT as described in Sec. \ref{sec: related_work}. It is essential to note that \citet{lagunas2021block} models (Hybrid and HybridNT) only prune the attention layers and not the entire model. To facilitate a comprehensive and standardized comparison of all algorithms, we recalibrated the size of their models based on our holistic perspective, ignoring any partial considerations. We combined the results obtained in their publication with those obtained from KEN and FLOP in Tab. \ref{tab: KEN vs Hybrid and other}. KEN outperformed all other compared models with a significant performance gap while utilizing fewer parameters in every instance. In addition to these findings, we conducted a thorough analysis of FLOP, which is the most complete pruning algorithm studied and, like KEN, decomposes original matrices to derive pruned ones. We conducted additional experiments on all models where FLOP could be applied, using the datasets listed in Tab. \ref{tab:Results}. We compared the results obtained from FLOP with those of KEN, which employed fewer parameters than FLOP. As shown in Tab. \ref{tab: KEN vs Flop}, FLOP outperforms KEN in only one instance. For all other models and datasets analyzed, KEN consistently outperforms FLOP.

\begin{table}[b!]
\centering
\begin{adjustbox}{width=\linewidth}
\begin{tabular}{lcc}
\hline
Model           & Trainable params                   & \begin{tabular}[c]{@{}c@{}}\texttt{glue-sst2} \\ Accuracy\end{tabular} \\ \hline
Bert-base       & 109M                   & 93.37              \\ \hline
Hybrid          & 94M          & 93.23          \\
HybridNT       & 94M          & 92.20          \\
\textbf{KEN}    & \textbf{80M}           & \textbf{93.80} \\ \hline
Hybrid         & 66M          & 91.97          \\
HybridNT       & 66M          & 90.71          \\
\citet{sajjad2020poor}          & 66M                    & 90.30           \\
\citet{gordon2020compressing}          & 66M                    & 90.80           \\
Flop            & 66M                    & 83.20           \\
\textbf{KEN}    & \textbf{63M}           & \textbf{92.90}  \\ \hline
\end{tabular}
\end{adjustbox}
\caption{Pruning algorithm comparations on SST-2 datasets}
\label{tab: KEN vs Hybrid and other}
\end{table}

\begin{table*}[htb!]
\centering
\begin{adjustbox}{width=\textwidth}
\begin{tabular}{lccccccc}
\hline
Model                    & \begin{tabular}[c]{@{}c@{}}Pruning\\ algorithm\end{tabular} & Trainable params. & \texttt{AG-NEWS} & \texttt{EMO} & \texttt{IMDB}  & \texttt{YELP\_POLARITY} & \texttt{glue-sst2} \\ \hline
\multirow{2}{*}{BLOOM$_{1B7}$} & KEN & 531M  & \textbf{92.2 ($\pm 0.1$)} & \textbf{90.6 ($\pm 0.1$)} & \textbf{84.2 ($\pm 0.1$)} & \textbf{96.3 ($\pm 0.1$)}  & \textbf{90.9 ($\pm 0.1$)}\\
& FLOP & 1.1B & 90.1 ($\pm 1.3$)& 84.0 ($\pm 1.9$) & 80.9 ($\pm 0.3$) & 85.5 ($\pm 3.5$) & 80.7 ($\pm 1.7$) \\ \hline
\multirow{2}{*}{BLOOM$_{560k}$} & KEN & 404M  & \textbf{91.3 ($\pm 0.1$)} & \textbf{85.5 ($\pm 3.5$)} & \textbf{81.3 ($\pm 0.3$)} & \textbf{94.8 ($\pm 0.5$)}  & \textbf{92.0 ($\pm 0.4$)}\\
& FLOP & 408M & 91.0 ($\pm 0.6$) & 84.0 ($\pm 2.3$) & 72.1 ($\pm 7.1$) & 87.0 ($\pm 0.5$) & 81.8 ($\pm 0.5$) \\ \hline
\multirow{2}{*}{DeBERTa} & KEN  & 84M & \textbf{91.4 ($\pm 0.6$)}& \textbf{88.9 ($\pm 1.5$)} & \textbf{82.5 ($\pm 3.1$)} & \textbf{96.0 ($\pm 0.2$)} &  \textbf{92.8 ($\pm 0.4$)} \\
& FLOP & 88M & 90.6 ($\pm 0.7$)& 83.1 ($\pm 1.7$) & 81.1 ($\pm 0.8$) & 91.4 ($\pm 0.1$) & 82.3 ($\pm 1.1$) \\ \hline
\multirow{2}{*}{Bert} & KEN & 57M  & \textbf{91.6 ($\pm 0.7$)} & \textbf{86.0 ($\pm 0.5$)} & \textbf{84.9 ($\pm 0.8$)} & \textbf{93.8 ($\pm 1.6$)}  & \textbf{92.8 ($\pm 0.5$)}\\
& FLOP & 66M & 90.9 ($\pm 0.9$)& 83.3 ($\pm 0.8$) & 80.5 ($\pm 0.6$)& 90.2 ($\pm 0.6$) & 83.2 ($\pm 0.2$) \\ \hline
\multirow{2}{*}{Ernie}& KEN  & 57M & \textbf{91.5 ($\pm 1.4$)} & \textbf{88.3 ($\pm 0.4$)} & \textbf{87.6 ($\pm 0.6$)} & \textbf{95.7 ($\pm 0.1$)} &      
\textbf{94.1 ($\pm 0.4$)}
\\
& FLOP & 67M & 89.8 ($\pm 0.4$ )& 83.8 ($\pm 2.3$) & 81.1 ($\pm 0.8$)& 90.9 ($\pm 0.1$) & 83.2 ($\pm 0.9$) \\ \hline
\multirow{2}{*}{DistilBERT} & KEN  & 40M & \textbf{91.9 ($\pm 0.3$)} & \textbf{88.2 ($\pm 1.1$)} & 78.1 ($\pm 1.4$)& \textbf{94.1 ($\pm 0.1$)} &  \textbf{89.2 ($\pm 0.7$)}\\
& FLOP & 45M & 90.7 ($\pm 0.9$) & 83.2 ($\pm 1.2$)&\textbf{81.2 ($\pm 0.9$)} & 90.7 ($\pm 0.1$)& 82.4 ($\pm 1.2$) \\ \hline
\multirow{2}{*}{Electra} & KEN                                                         & 14M  & \textbf{91.3 ($\pm 0.2$)} & \textbf{85.6 ($\pm 0.3$)} & \textbf{84.3 ($\pm 0.1$)} & \textbf{93.7 ($\pm 0.4$)} &  \textbf{90.1 ($\pm 0.1$)} \\
& FLOP & 28M & 90.9 ($\pm 0.3$) & 83.1 ($\pm 2.1$) & 81.2 ($\pm 0.1$) & 90.5 ($\pm 0.1$) & 81.1 ($\pm 0.3$) \\ \hline
\end{tabular}
\end{adjustbox}
\caption{Comparation between KEN and FLOP pruning algorithms on different datasets. Mean and standard deviation are calculated on equal runs for each dataset and algorithm analyzed. The \textit{Trainable params} column indicates the number of parameters used by each algorithm after the pruning phase.}

\label{tab: KEN vs Flop}
\end{table*}

\begin{figure*}[h!]
    \centering
    \includegraphics[width=\linewidth]{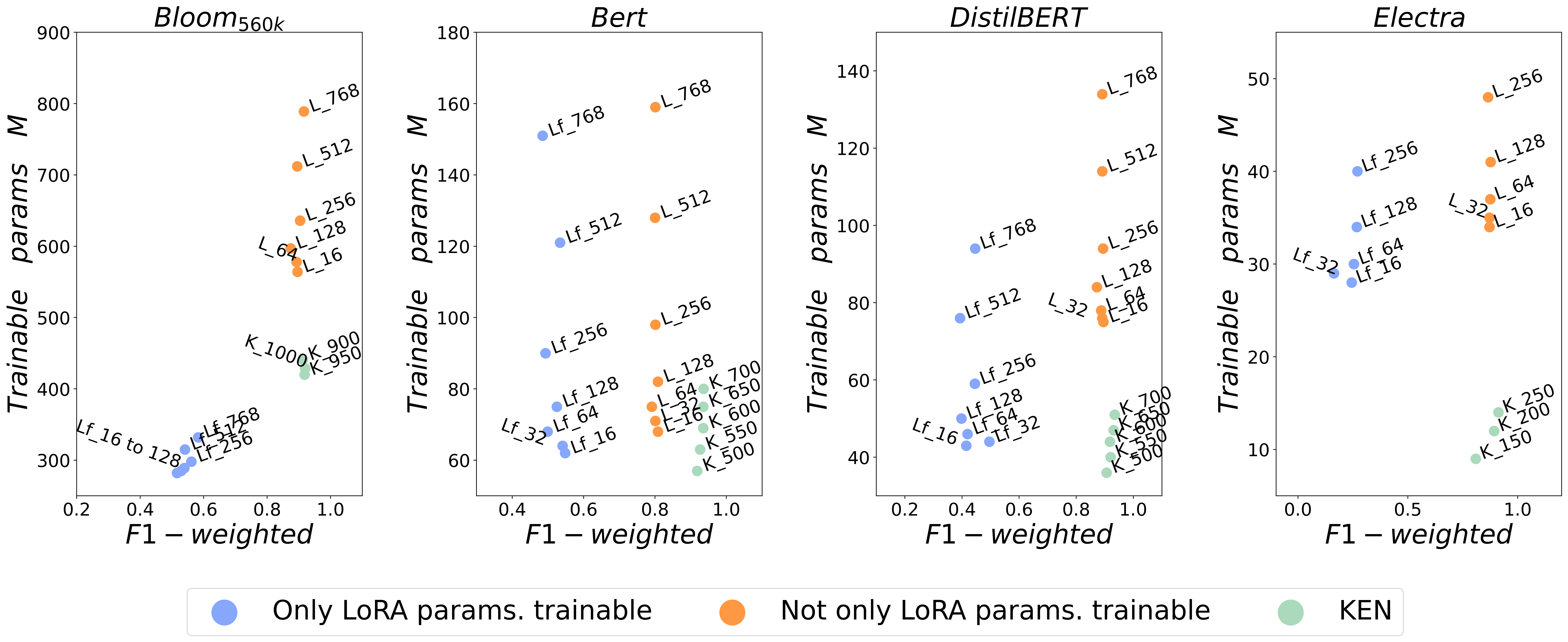}
    \caption{Comparison between KEN and LoRA. Labels for the LoRA marker indicate the dimension of the rank-decomposition matrix analyzed while, for KEN, the $k$ value used}
    \label{fig: Lora experiments}
\end{figure*}

Although KEN belongs to the \textit{winning ticket pruning} algorithms family, it shares similarities with Parameter Efficient Fine-tuning (PEFT) algorithms. This is because both approaches aim to identify a subset of optimal parameters within the fine-tuned model. We thoroughly evaluated KEN and compared it to LoRA, which is currently the state-of-the-art PEFT algorithm. We applied LoRA and KEN to the same layers of each model and then trained the LoRA-based models using five times more training epochs than their KEN-based counterparts.
Additionally, we gradually increased the number of rank decomposition matrices for each model from 16 to 768, which is the average size of the matrices in the tested models. In each LoRA-based experiment, only the LoRA-specific parameters were designated as either trainable or not. Our results, presented in Fig. \ref{fig: Lora experiments}, demonstrate that KEN consistently outperforms LoRA in terms of F1-measure while utilizing fewer trained parameters. However, when LoRA parameters are not the only ones trained, KEN and LoRA generally produce similar results. Nevertheless, LoRA consistently requires a larger parameter count than KEN. These compelling results provide strong evidence supporting our hypothesis that strategically selecting a subset of parameters and resetting the remainder offers a promising alternative to conventional pruning techniques.

\subsection{Compression values} \label{subsec: compression results}

One of KEN main goals, is to significantly reduce the overall size of transformer models, including their file sizes. To achieve this, KEN leverages a subnetwork comprising only $k$-trained parameters, allowing it to be saved and then injected into its pre-trained counterpart. This process requires a support file, like a dictionary, that specifies the precise location of each saved parameter within the pre-trained model. To guarantee an unbiased comparison between the original and compressed model sizes, the compressed one is saved using identical techniques and format as the original model.  For each transformer analyzed, two compressed versions are generated using both high and low $k$ values.

\begin{table}[h!]
\centering
\begin{adjustbox}{width=\linewidth}
\begin{tabular}{|c|c|c|c|r|}
\hline
\multirow{2}{*}{Model} &
  \multirow{2}{*}{\begin{tabular}[c]{@{}c@{}}Total \\ params\end{tabular}} &
  \multirow{2}{*}{\begin{tabular}[c]{@{}c@{}}Original \\ file size\end{tabular}} &
  \multirow{2}{*}{\begin{tabular}[c]{@{}c@{}}\# trainable \\ params\end{tabular}} &
  \multirow{2}{*}{\begin{tabular}[c]{@{}c@{}}Compressed file size \\ (Model + support dict)\end{tabular}} \\
                            &                       &                         &      &                        \\ \hline
\multirow{2}{*}{BLOOM$_{1B7}$}       & \multirow{2}{*}{1.72B} & \multirow{2}{*}{7,055 MB} & 664M  & \textbf{3,071 MB} (2,923 + 148)      \\ \cline{4-5} 
                            &                       &                         & 442M  & \textbf{2137 MB} (2,013 + 124)  \\ \hline
\multirow{2}{*}{BLOOM$_{560k}$}       & \multirow{2}{*}{560M} & \multirow{2}{*}{2,294 MB} & 429M  & \textbf{2,084 MB} (1,956 + 128)      \\ \cline{4-5} 
                            &                       &                         & 386M  & \textbf{1,842 MB} (1,731 + 111)  \\ \hline
\multirow{2}{*}{BERT}       & \multirow{2}{*}{109M} & \multirow{2}{*}{438 MB} & 80M  & \textbf{358 MB} (320 + 38)      \\ \cline{4-5} 
                            &                       &                         & 57M  & \textbf{260.2 MB} (228 + 32.2)  \\ \hline
\multirow{2}{*}{DistilBERT} & \multirow{2}{*}{66M}  & \multirow{2}{*}{266 MB} & 51M  & \textbf{231.4 MB} (203 + 28.4)  \\ \cline{4-5} 
                            &                       &                         & 36M  & \textbf{165 MB} (145 + 20)      \\ \hline
\multirow{2}{*}{DeBERTa}    & \multirow{2}{*}{138M} & \multirow{2}{*}{555 MB} & 107M & \textbf{476.3 MB} (428 + 48.3)  \\ \cline{4-5} 
                            &                       &                         & 76M  & \textbf{348.4 MB} (306 + 42.4)  \\ \hline
\multirow{2}{*}{Ernie}      & \multirow{2}{*}{109M} & \multirow{2}{*}{438 MB} & 80M  & \textbf{356.9 MB} (320 + 36.9)  \\ \cline{4-5} 
                            &                       &                         & 57M  & \textbf{260.3 MB} (228 + 32.3)  \\ \hline
\multirow{2}{*}{Electra}    & \multirow{2}{*}{33M}  & \multirow{2}{*}{134 MB} & 14M  & \textbf{67.01 MB} (59.1 + 7.91) \\ \cline{4-5} 
                            &                       &                         & 9M   & \textbf{42.58 MB} (35.5 + 7.08) \\ \hline
\end{tabular}
\end{adjustbox}
\caption{Comparison of the .pt file size between the original and compressed transformer weights}
\label{tab:Compression_table}
\end{table}

As shown in Tab. \ref{tab:Compression_table}, both versions of the compressed models demonstrate substantial memory savings, directly proportional to the number of saved parameters. Specifically, transformers saved using a high $k$ value, thus closely mirroring the structure of the unpruned model, conserving $\geq \approx $100 MB per each. This value increases further as the number of trained parameters saved diminishes. The support dictionary for parameter injection, stored using the Lempel-Ziv-Markov chain data compression algorithm, has a negligible impact on the final model weight, which remains significantly smaller than the original. Additionally, the time required to load the injected parameters into the pre-trained model scales linearly with the transformer architecture and the compression employed.

\subsection{KEN$_{viz}$} \label{subsec:KEN_viz}
\begin{figure}[h!]
     \centering
     \begin{subfigure}[b]{0.49\linewidth}
         \centering
         \includegraphics[width=\linewidth]{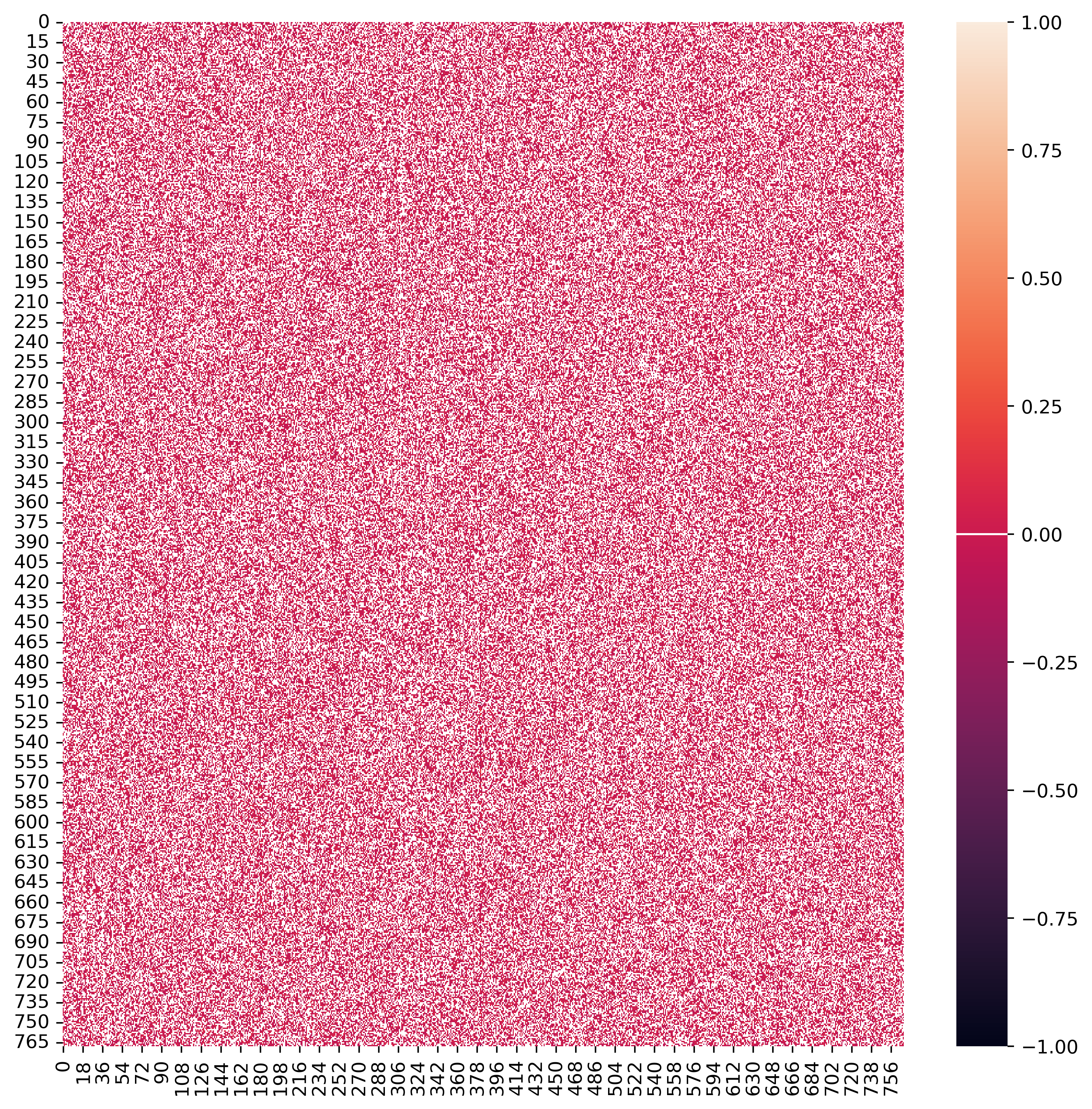}
         \caption{Single Matrix View}
         \label{fig:Kenviz optimized}
     \end{subfigure}
     \hfill
     \begin{subfigure}[b]{0.49\linewidth}
         \centering
         \includegraphics[width=\linewidth]{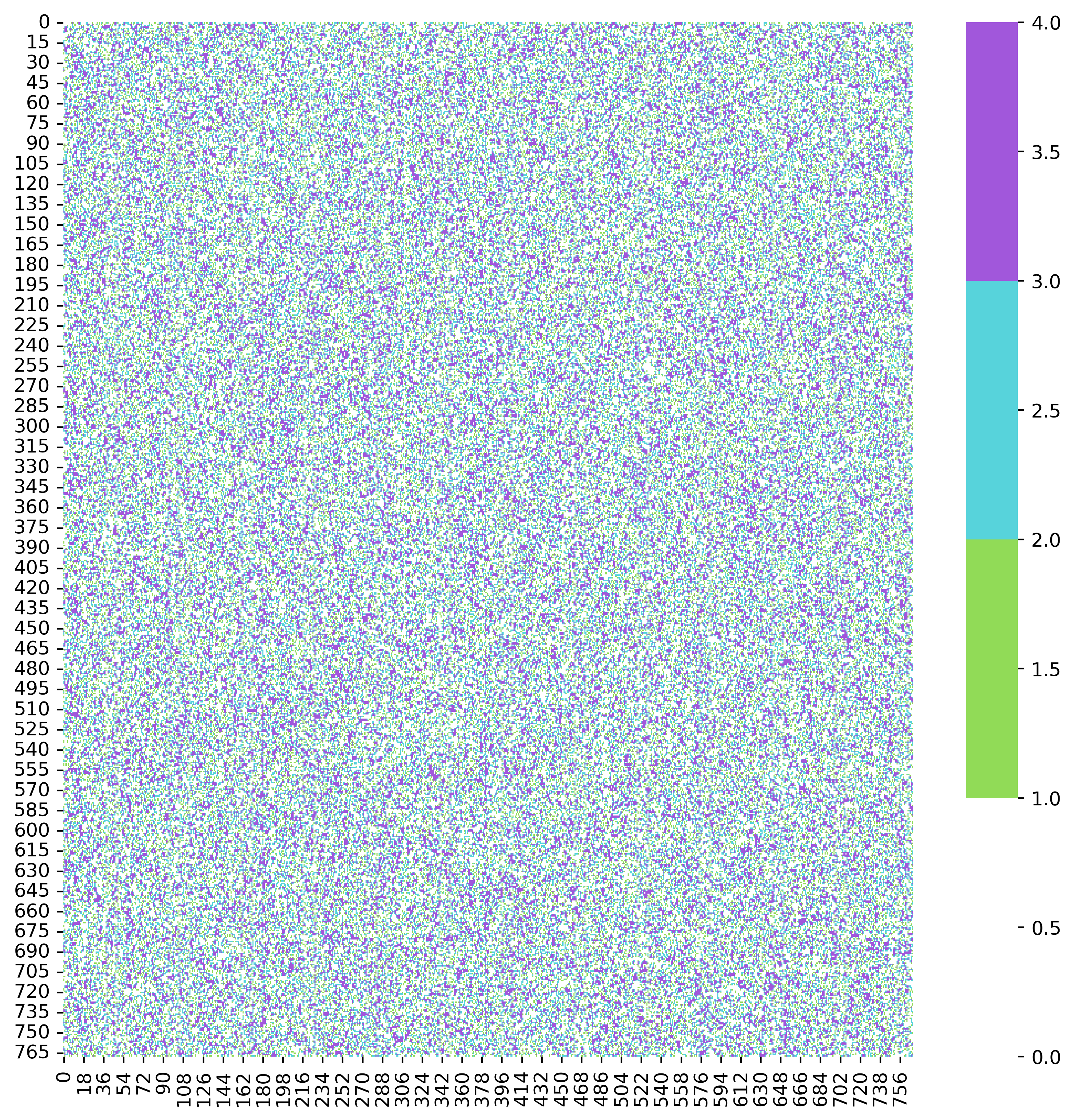}
         \caption{Neighbor Count View}
         \label{fig:Kenviz cluster}
     \end{subfigure}
    \caption{Output of KEN$_{viz}$ of the key attention matrix at layer 12 of a BERT model trained on \texttt{glue-sst2}. Reset parameters 47.92\%}
    \label{fig: Neighbor Count example}
\end{figure}

KEN$_{viz}$ is a visualization tool that provides a clear understanding of matrices composition after the application of KEN pruning step. It offers various views to explore the pruned model, including:
\begin{enumerate}
    \item \textbf{Single Matrix View}: This view offers a clear understanding of the parameters retained by KEN, leaving the pruned elements blank (Fig. \ref{fig: Sparse carosel}, Fig. \ref{fig:Kenviz optimized}).
    
    \item \textbf{Neighbor Count View}: It visualizes the number of non-zero neighbors (horizontally and vertically) for each point in a given matrix. This provides additional information about the remaining parameters and potential parameter clusters that might emerge (Fig. \ref{fig:Kenviz cluster}).
    
    \item  \textbf{Layer-wise View}: This view shows the impact of KEN on the entire model architecture. It iteratively applies the \textit{Single matrix view} to the same matrix on all layers it appears. This allows a layer-by-layer comparison, revealing how the pruning of the same matrix changes in different parts of the network.
\end{enumerate}

The examples in Fig. \ref{fig: Neighbor Count example} and Apx. \ref{Apx: KEN_viz} both indicate that the number of non-zero neighbors for each point remains consistently high even in cases with high reset parameters. This suggests that the chosen parameters not only represent the most effective elements but also display a well-proportioned distribution within each matrix.

\section{Conclusions}
This paper introduces KEN: a novel non-architecture-specific pruning algorithm that leverages KDE to construct an abstraction of the parameter distribution and selectively retain a finite subset of them while resetting the others to their pre-trained states. Our extensive evaluations on seven different transformer models demonstrate that KEN consistently achieves remarkable compression rates, reducing unnecessary parameters by a minimum of 25\% up to $\approx$ 70\% on some models, without compromising model performance. Moreover, by leveraging the KEN core idea, is possible to store only the active subnetwork, leading to substantial memory savings.

We also present KEN$_{viz}$: the KEN visualizer that provides insights into the algorithm operation. KEN$_{viz}$ reveals that KEN uniformly selects parameters across matrices, preventing parameter clusters. With KEN we demonstrate how a simple, non-parametric algorithm commonly used in statistics, can be effectively adopted for model pruning achieving excellent results in both compression and performance.

\section{Limitations}
One of the major limitations of KEN is its computational efficiency, especially when analyzing large models. Although KEN excels at generating rich distributions with high $k$ values, it faces a trade-off in computational efficiency, particularly for large models. Processing time scales linearly with the model size, number of layers, and chosen $k$. This primarily impacts the parameter selection stage, not compressed model storage or loading.

Our focus on the sequence classification task ensured consistent results, but unpublished experiments suggest KEN effectiveness extends to other tasks. Future work will explore this broader applicability and investigate potential optimizations for large-scale deployments.

\newpage
\bibliography{anthology,custom}

\begin{thebibliography}{50}
\expandafter\ifx\csname natexlab\endcsname\relax\def\natexlab#1{#1}\fi

\bibitem[{Ba and Caruana(2014)}]{ba2014deep}
Jimmy Ba and Rich Caruana. 2014.
\newblock Do deep nets really need to be deep?
\newblock \emph{Advances in neural information processing systems}, 27.

\bibitem[{Barbieri et~al.(2020)Barbieri, Camacho-Collados, Espinosa~Anke, and Neves}]{barbieri-etal-2020-tweeteval}
Francesco Barbieri, Jose Camacho-Collados, Luis Espinosa~Anke, and Leonardo Neves. 2020.
\newblock \href {https://doi.org/10.18653/v1/2020.findings-emnlp.148} {{T}weet{E}val: Unified benchmark and comparative evaluation for tweet classification}.
\newblock In \emph{Findings of the Association for Computational Linguistics: EMNLP 2020}, pages 1644--1650, Online. Association for Computational Linguistics.

\bibitem[{Bastings et~al.(2019)Bastings, Aziz, and Titov}]{bastings2019interpretable}
Jasmijn Bastings, Wilker Aziz, and Ivan Titov. 2019.
\newblock Interpretable neural predictions with differentiable binary variables.
\newblock \emph{arXiv preprint arXiv:1905.08160}.

\bibitem[{Benbaki et~al.(2023)Benbaki, Chen, Meng, Hazimeh, Ponomareva, Zhao, and Mazumder}]{benbaki2023fast}
Riade Benbaki, Wenyu Chen, Xiang Meng, Hussein Hazimeh, Natalia Ponomareva, Zhe Zhao, and Rahul Mazumder. 2023.
\newblock Fast as chita: Neural network pruning with combinatorial optimization.
\newblock \emph{arXiv preprint arXiv:2302.14623}.

\bibitem[{Blalock et~al.(2020)Blalock, Gonzalez~Ortiz, Frankle, and Guttag}]{blalock2020state}
Davis Blalock, Jose~Javier Gonzalez~Ortiz, Jonathan Frankle, and John Guttag. 2020.
\newblock What is the state of neural network pruning?
\newblock \emph{Proceedings of machine learning and systems}, 2:129--146.

\bibitem[{Chatterjee et~al.(2019)Chatterjee, Narahari, Joshi, and Agrawal}]{chatterjee-etal-2019-semeval}
Ankush Chatterjee, Kedhar~Nath Narahari, Meghana Joshi, and Puneet Agrawal. 2019.
\newblock \href {https://doi.org/10.18653/v1/S19-2005} {{S}em{E}val-2019 task 3: {E}mo{C}ontext contextual emotion detection in text}.
\newblock In \emph{Proceedings of the 13th International Workshop on Semantic Evaluation}, pages 39--48, Minneapolis, Minnesota, USA. Association for Computational Linguistics.

\bibitem[{Clark et~al.(2020)Clark, Luong, Le, and Manning}]{clark2020electra}
Kevin Clark, Minh-Thang Luong, Quoc~V Le, and Christopher~D Manning. 2020.
\newblock Electra: Pre-training text encoders as discriminators rather than generators.
\newblock \emph{arXiv preprint arXiv:2003.10555}.

\bibitem[{Cohan et~al.(2019)Cohan, Ammar, van Zuylen, and Cady}]{cohan-etal-2019-structural}
Arman Cohan, Waleed Ammar, Madeleine van Zuylen, and Field Cady. 2019.
\newblock \href {https://doi.org/10.18653/v1/N19-1361} {Structural scaffolds for citation intent classification in scientific publications}.
\newblock In \emph{Proceedings of the 2019 Conference of the North {A}merican Chapter of the Association for Computational Linguistics: Human Language Technologies, Volume 1 (Long and Short Papers)}, pages 3586--3596, Minneapolis, Minnesota. Association for Computational Linguistics.

\bibitem[{Davidson et~al.(2017)Davidson, Warmsley, Macy, and Weber}]{davidson2017automated}
Thomas Davidson, Dana Warmsley, Michael Macy, and Ingmar Weber. 2017.
\newblock Automated hate speech detection and the problem of offensive language.
\newblock In \emph{Proceedings of the international AAAI conference on web and social media}, volume~11, pages 512--515.

\bibitem[{de~Gibert et~al.(2018)de~Gibert, Perez, Garc{\'\i}a-Pablos, and Cuadros}]{gibert2018hate}
Ona de~Gibert, Naiara Perez, Aitor Garc{\'\i}a-Pablos, and Montse Cuadros. 2018.
\newblock \href {https://doi.org/10.18653/v1/W18-5102} {{Hate Speech Dataset from a White Supremacy Forum}}.
\newblock In \emph{Proceedings of the 2nd Workshop on Abusive Language Online ({ALW}2)}, pages 11--20, Brussels, Belgium. Association for Computational Linguistics.

\bibitem[{Devlin et~al.(2018)Devlin, Chang, Lee, and Toutanova}]{devlin2018bert}
Jacob Devlin, Ming-Wei Chang, Kenton Lee, and Kristina Toutanova. 2018.
\newblock Bert: Pre-training of deep bidirectional transformers for language understanding.
\newblock \emph{arXiv preprint arXiv:1810.04805}.

\bibitem[{Dong et~al.(2017)Dong, Chen, and Pan}]{dong2017learning}
Xin Dong, Shangyu Chen, and Sinno Pan. 2017.
\newblock Learning to prune deep neural networks via layer-wise optimal brain surgeon.
\newblock \emph{Advances in neural information processing systems}, 30.

\bibitem[{Frankle and Carbin(2018)}]{frankle2018lottery}
Jonathan Frankle and Michael Carbin. 2018.
\newblock The lottery ticket hypothesis: Finding sparse, trainable neural networks.
\newblock \emph{arXiv preprint arXiv:1803.03635}.

\bibitem[{Gong et~al.(2014)Gong, Liu, Yang, and Bourdev}]{gong2014compressing}
Yunchao Gong, Liu Liu, Ming Yang, and Lubomir Bourdev. 2014.
\newblock Compressing deep convolutional networks using vector quantization.
\newblock \emph{arXiv preprint arXiv:1412.6115}.

\bibitem[{Gordon et~al.(2020)Gordon, Duh, and Andrews}]{gordon2020compressing}
Mitchell~A Gordon, Kevin Duh, and Nicholas Andrews. 2020.
\newblock Compressing bert: Studying the effects of weight pruning on transfer learning.
\newblock \emph{arXiv preprint arXiv:2002.08307}.

\bibitem[{Gulli(2005)}]{gulliAgnews}
Antonio Gulli. 2005.
\newblock \href {http://groups.di.unipi.it/~gulli/AG_corpus_of_news_articles.html} {Ag's corpus of news articles}.

\bibitem[{Gurulingappa et~al.(2012)Gurulingappa, Rajput, Roberts, Fluck, Hofmann-Apitius, and Toldo}]{ade}
Harsha Gurulingappa, Abdul~Mateen Rajput, Angus Roberts, Juliane Fluck, Martin Hofmann-Apitius, and Luca Toldo. 2012.
\newblock \href {https://doi.org/https://doi.org/10.1016/j.jbi.2012.04.008} {Development of a benchmark corpus to support the automatic extraction of drug-related adverse effects from medical case reports}.
\newblock \emph{Journal of Biomedical Informatics}, 45(5):885 -- 892.
\newblock Text Mining and Natural Language Processing in Pharmacogenomics.

\bibitem[{Han et~al.(2015)Han, Pool, Tran, and Dally}]{han2015learning}
Song Han, Jeff Pool, John Tran, and William Dally. 2015.
\newblock Learning both weights and connections for efficient neural network.
\newblock \emph{Advances in neural information processing systems}, 28.

\bibitem[{Hanson and Pratt(1988)}]{hanson1988comparing}
Stephen Hanson and Lorien Pratt. 1988.
\newblock Comparing biases for minimal network construction with back-propagation.
\newblock \emph{Advances in neural information processing systems}, 1.

\bibitem[{Hassibi and Stork(1992)}]{hassibi1992second}
Babak Hassibi and David Stork. 1992.
\newblock Second order derivatives for network pruning: Optimal brain surgeon.
\newblock \emph{Advances in neural information processing systems}, 5.

\bibitem[{He et~al.(2020)He, Liu, Gao, and Chen}]{he2020deberta}
Pengcheng He, Xiaodong Liu, Jianfeng Gao, and Weizhu Chen. 2020.
\newblock Deberta: Decoding-enhanced bert with disentangled attention.
\newblock \emph{arXiv preprint arXiv:2006.03654}.

\bibitem[{Hu et~al.(2021)Hu, Shen, Wallis, Allen-Zhu, Li, Wang, Wang, and Chen}]{hu2021lora}
Edward~J Hu, Yelong Shen, Phillip Wallis, Zeyuan Allen-Zhu, Yuanzhi Li, Shean Wang, Lu~Wang, and Weizhu Chen. 2021.
\newblock Lora: Low-rank adaptation of large language models.
\newblock \emph{arXiv preprint arXiv:2106.09685}.

\bibitem[{Huang et~al.(2018)Huang, Liu, Van~der Maaten, and Weinberger}]{huang2018condensenet}
Gao Huang, Shichen Liu, Laurens Van~der Maaten, and Kilian~Q Weinberger. 2018.
\newblock Condensenet: An efficient densenet using learned group convolutions.
\newblock In \emph{Proceedings of the IEEE conference on computer vision and pattern recognition}, pages 2752--2761.

\bibitem[{Janowsky(1989)}]{janowsky1989pruning}
Steven~A Janowsky. 1989.
\newblock Pruning versus clipping in neural networks.
\newblock \emph{Physical Review A}, 39(12):6600.

\bibitem[{Joulin et~al.(2017)Joulin, Ciss{\'e}, Grangier, J{\'e}gou et~al.}]{joulin2017efficient}
Armand Joulin, Moustapha Ciss{\'e}, David Grangier, Herv{\'e} J{\'e}gou, et~al. 2017.
\newblock Efficient softmax approximation for gpus.
\newblock In \emph{International conference on machine learning}, pages 1302--1310. PMLR.

\bibitem[{Keung et~al.(2020)Keung, Lu, Szarvas, and Smith}]{marc_reviews}
Phillip Keung, Yichao Lu, György Szarvas, and Noah~A. Smith. 2020.
\newblock The multilingual amazon reviews corpus.
\newblock In \emph{Proceedings of the 2020 Conference on Empirical Methods in Natural Language Processing}.

\bibitem[{Kim and Rush(2016)}]{kim2016sequence}
Yoon Kim and Alexander~M Rush. 2016.
\newblock Sequence-level knowledge distillation.
\newblock \emph{arXiv preprint arXiv:1606.07947}.

\bibitem[{Lagunas et~al.(2021)Lagunas, Charlaix, Sanh, and Rush}]{lagunas2021block}
Fran{\c{c}}ois Lagunas, Ella Charlaix, Victor Sanh, and Alexander~M Rush. 2021.
\newblock Block pruning for faster transformers.
\newblock \emph{arXiv preprint arXiv:2109.04838}.

\bibitem[{LeCun et~al.(1989)LeCun, Denker, and Solla}]{lecun1989optimal}
Yann LeCun, John Denker, and Sara Solla. 1989.
\newblock Optimal brain damage.
\newblock \emph{Advances in neural information processing systems}, 2.

\bibitem[{Li and Roth(2002)}]{li-roth-2002-learning}
Xin Li and Dan Roth. 2002.
\newblock \href {https://www.aclweb.org/anthology/C02-1150} {Learning question classifiers}.
\newblock In \emph{{COLING} 2002: The 19th International Conference on Computational Linguistics}.

\bibitem[{Maas et~al.(2011)Maas, Daly, Pham, Huang, Ng, and Potts}]{maas-etal-2011-learning}
Andrew~L. Maas, Raymond~E. Daly, Peter~T. Pham, Dan Huang, Andrew~Y. Ng, and Christopher Potts. 2011.
\newblock \href {https://aclanthology.org/P11-1015} {Learning word vectors for sentiment analysis}.
\newblock In \emph{Proceedings of the 49th Annual Meeting of the Association for Computational Linguistics: Human Language Technologies}, pages 142--150, Portland, Oregon, USA. Association for Computational Linguistics.

\bibitem[{Malach et~al.(2020)Malach, Yehudai, Shalev-Schwartz, and Shamir}]{malach2020proving}
Eran Malach, Gilad Yehudai, Shai Shalev-Schwartz, and Ohad Shamir. 2020.
\newblock Proving the lottery ticket hypothesis: Pruning is all you need.
\newblock In \emph{International Conference on Machine Learning}, pages 6682--6691. PMLR.

\bibitem[{Mozer and Smolensky(1989)}]{mozer1989using}
Michael~C Mozer and Paul Smolensky. 1989.
\newblock Using relevance to reduce network size automatically.
\newblock \emph{Connection Science}, 1(1):3--16.

\bibitem[{Pang and Lee(2005)}]{pang2005seeing}
Bo~Pang and Lillian Lee. 2005.
\newblock Seeing stars: Exploiting class relationships for sentiment categorization with respect to rating scales.
\newblock \emph{arXiv preprint cs/0506075}.

\bibitem[{Sajjad et~al.(2020)Sajjad, Dalvi, Durrani, and Nakov}]{sajjad2020poor}
Hassan Sajjad, Fahim Dalvi, Nadir Durrani, and Preslav Nakov. 2020.
\newblock Poor man’s bert: Smaller and faster transformer models.
\newblock \emph{arXiv preprint arXiv:2004.03844}, 2(2).

\bibitem[{Sanh et~al.(2019)Sanh, Debut, Chaumond, and Wolf}]{sanh2019distilbert}
Victor Sanh, Lysandre Debut, Julien Chaumond, and Thomas Wolf. 2019.
\newblock Distilbert, a distilled version of bert: smaller, faster, cheaper and lighter.
\newblock \emph{arXiv preprint arXiv:1910.01108}.

\bibitem[{Sanh et~al.(2020)Sanh, Wolf, and Rush}]{sanh2020movement}
Victor Sanh, Thomas Wolf, and Alexander Rush. 2020.
\newblock Movement pruning: Adaptive sparsity by fine-tuning.
\newblock \emph{Advances in Neural Information Processing Systems}, 33:20378--20389.

\bibitem[{Scott(2015)}]{scott2015multivariate}
David~W Scott. 2015.
\newblock \emph{Multivariate density estimation: theory, practice, and visualization}.
\newblock John Wiley \& Sons.

\bibitem[{Sheng and Uthus(2020)}]{sheng-uthus-2020-investigating}
Emily Sheng and David Uthus. 2020.
\newblock \href {https://aclanthology.org/2020.gebnlp-1.9} {Investigating societal biases in a poetry composition system}.
\newblock In \emph{Proceedings of the Second Workshop on Gender Bias in Natural Language Processing}, pages 93--106, Barcelona, Spain (Online). Association for Computational Linguistics.

\bibitem[{Singh and Alistarh(2020)}]{singh2020woodfisher}
Sidak~Pal Singh and Dan Alistarh. 2020.
\newblock Woodfisher: Efficient second-order approximation for neural network compression.
\newblock \emph{Advances in Neural Information Processing Systems}, 33:18098--18109.

\bibitem[{Socher et~al.(2013)Socher, Perelygin, Wu, Chuang, Manning, Ng, and Potts}]{socher-etal-2013-recursive}
Richard Socher, Alex Perelygin, Jean Wu, Jason Chuang, Christopher~D. Manning, Andrew Ng, and Christopher Potts. 2013.
\newblock \href {https://aclanthology.org/D13-1170} {Recursive deep models for semantic compositionality over a sentiment treebank}.
\newblock In \emph{Proceedings of the 2013 Conference on Empirical Methods in Natural Language Processing}, pages 1631--1642, Seattle, Washington, USA. Association for Computational Linguistics.

\bibitem[{Sun et~al.(2020)Sun, Wang, Li, Feng, Tian, Wu, and Wang}]{sun2020ernie}
Yu~Sun, Shuohuan Wang, Yukun Li, Shikun Feng, Hao Tian, Hua Wu, and Haifeng Wang. 2020.
\newblock Ernie 2.0: A continual pre-training framework for language understanding.
\newblock In \emph{Proceedings of the AAAI conference on artificial intelligence}, volume~34, pages 8968--8975.

\bibitem[{Sze et~al.(2017)Sze, Chen, Yang, and Emer}]{sze2017efficient}
Vivienne Sze, Yu-Hsin Chen, Tien-Ju Yang, and Joel~S Emer. 2017.
\newblock Efficient processing of deep neural networks: A tutorial and survey.
\newblock \emph{Proceedings of the IEEE}, 105(12):2295--2329.

\bibitem[{Vaswani et~al.(2017)Vaswani, Shazeer, Parmar, Uszkoreit, Jones, Gomez, Kaiser, and Polosukhin}]{vaswani2017attention}
Ashish Vaswani, Noam Shazeer, Niki Parmar, Jakob Uszkoreit, Llion Jones, Aidan~N Gomez, {\L}ukasz Kaiser, and Illia Polosukhin. 2017.
\newblock Attention is all you need.
\newblock \emph{Advances in neural information processing systems}, 30.

\bibitem[{Wang et~al.(2019)Wang, Wohlwend, and Lei}]{wang2019structured}
Ziheng Wang, Jeremy Wohlwend, and Tao Lei. 2019.
\newblock Structured pruning of large language models.
\newblock \emph{arXiv preprint arXiv:1910.04732}.

\bibitem[{Workshop et~al.(2022)Workshop, Scao, Fan, Akiki, Pavlick, Ili{\'c}, Hesslow, Castagn{\'e}, Luccioni, Yvon et~al.}]{workshop2022bloom}
BigScience Workshop, Teven~Le Scao, Angela Fan, Christopher Akiki, Ellie Pavlick, Suzana Ili{\'c}, Daniel Hesslow, Roman Castagn{\'e}, Alexandra~Sasha Luccioni, Fran{\c{c}}ois Yvon, et~al. 2022.
\newblock Bloom: A 176b-parameter open-access multilingual language model.
\newblock \emph{arXiv preprint arXiv:2211.05100}.

\bibitem[{Yang et~al.(2017)Yang, Chen, and Sze}]{yang2017designing}
Tien-Ju Yang, Yu-Hsin Chen, and Vivienne Sze. 2017.
\newblock Designing energy-efficient convolutional neural networks using energy-aware pruning.
\newblock In \emph{Proceedings of the IEEE conference on computer vision and pattern recognition}, pages 5687--5695.

\bibitem[{Zhang et~al.(2015)Zhang, Zhao, and LeCun}]{zhang2015character}
Xiang Zhang, Junbo Zhao, and Yann LeCun. 2015.
\newblock Character-level convolutional networks for text classification.
\newblock \emph{Advances in neural information processing systems}, 28.

\bibitem[{Zhu et~al.(2016)Zhu, Han, Mao, and Dally}]{zhu2016trained}
Chenzhuo Zhu, Song Han, Huizi Mao, and William~J Dally. 2016.
\newblock Trained ternary quantization.
\newblock \emph{arXiv preprint arXiv:1612.01064}.

\bibitem[{Zhu and Gupta(2017)}]{zhu2017prune}
Michael Zhu and Suyog Gupta. 2017.
\newblock To prune, or not to prune: exploring the efficacy of pruning for model compression.
\newblock \emph{arXiv preprint arXiv:1710.01878}.

\end{thebibliography}
\bibliographystyle{acl_natbib}

\begin{figure*}[h!]
    \centering
    \includegraphics[width=\linewidth]{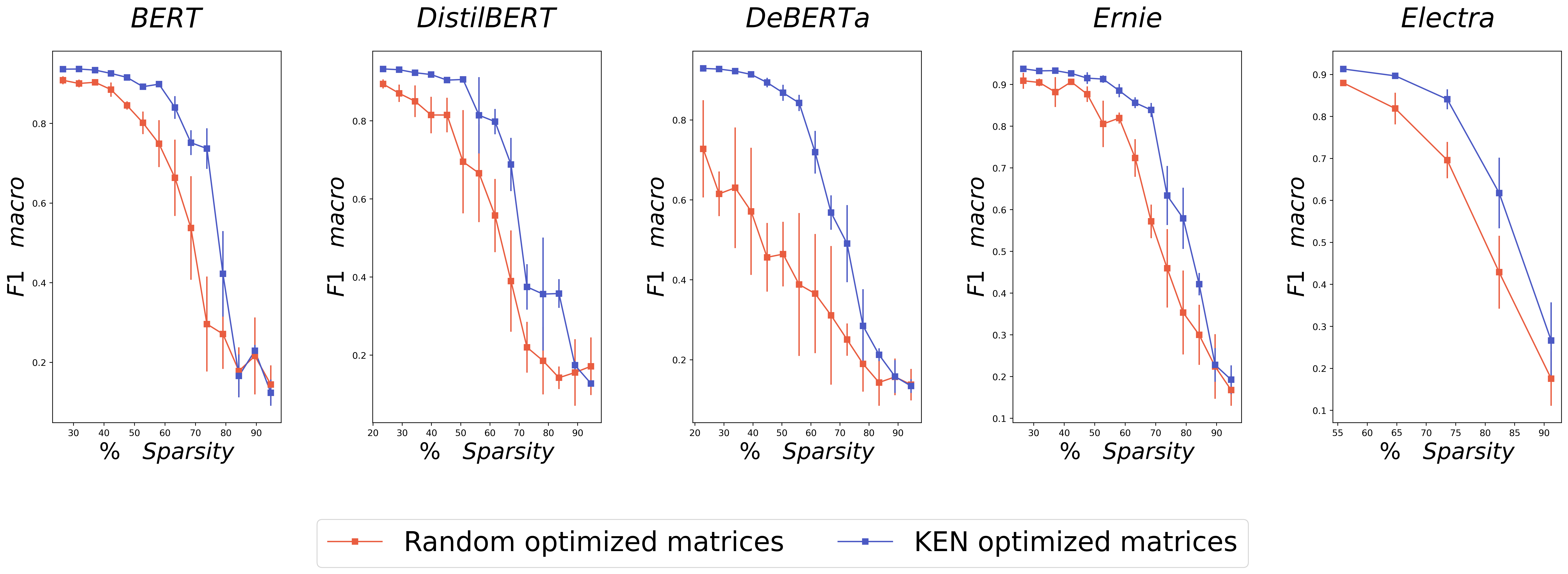}
    \caption{Performance variation on \texttt{AG-NEWS} dataset with different reset parameters percentage value. All the experiments were conducted using KEN \textit{full} configuration}
    \label{fig: injected parameters}
\end{figure*}

\newpage
\appendix

\section{How to prove the importance of selected parameters} \label{Appendix: random KEN}

To assess the effectiveness of KEN core idea, which involves selecting parameters based on their distribution using Kernel Density Estimation (KDE), we conducted parallel experiments. In these experiments, we compared KEN against random parameter pruning. Random parameters were either retained or reset to pre-trained values. This enabled us to determine if the selection method used in KEN resulted in a better subnetwork compared to random selection

Formally, for each matrix in a generic model, the optimized matrix $\hat{W}$ contained $k$ randomly selected fine-tuned parameters. Our goal is to determine whether the parameters introduced into a generic transformer model by KEN constituted an optimal subnetwork or if equivalent results could be achieved by randomly selecting the same number of parameters. To address this question, we performed an experiment using the \texttt{AG-NEWS} dataset, comparing the performance differences between extracting $\hat{W}$ matrices using KEN and using $k$ random values for each matrix row.

The results, illustrated in Fig. \ref{fig: injected parameters}, consistently show that KEN outperforms its random counterpart. KEN achieves a lower error rate and maintains higher performance at reasonable compression levels. However, for all models tested, a compression threshold exists beyond which performance inevitably declines. The KEN algorithm effectively compresses models while minimizing this decline, preserving high performance. Conversely, random parameter selection reaches this threshold earlier, resulting in a larger performance drop and higher error rate. Notably, the best achievable performance with random selection is always lower than or equal to the average achieved with KEN.

Furthermore, when using KEN, the error rate remains minimal within the threshold. This suggests that the subnetwork derived from KEN is not random; rather, it consistently selects the most effective portion of the original network.

\begin{table*}[h!]
\centering
\begin{adjustbox}{width=\linewidth}
\begin{tabular}{lccccccc}
Dataset                                 &BLOOM$_{1B7}$& BLOOM$_{560k}$ & Bert       & DistilBert   & DeBERTa       & Ernie      & Electra      \\ \hline
\texttt{trec}                           & 61.46\% & 23.26\% & 26.55\%    & 23.45\%      & 22.84\%       & 26.55\%    & 55.94\%      \\
\texttt{rotten\_tomatoes}               & 69.17\% & 24.80\% & 26.55\%    & 34.39\%      & 44.88\%       & 42.29\%    & 55.94\%      \\
\texttt{hate\_speech\_offensive}        & 61.46\% & 23.26\% & 26.55\%    & 34.39\%      & 22.84\%       & 26.55\%    & 55.94\%      \\
\texttt{hate\_speech18}                 & 61.46\% & 23.26\% & 26.55\%    & 23.45\%      & 33.86\%       & 31.80\%    & 64.75\%      \\
\texttt{scicite}                        & 61.46\% & 23.26\% & 37.05\%    & 28.92\%      & 22.84\%       & 31.80\%    & 55.94\%$^{\dagger}$   \\
\texttt{ade\_corpus\_v2}                & 69.17\% & 24.80\% & 52.78\%    & 45.32\%      & 44.88\%       & 63.28\%    & 73.56\%      \\
\texttt{amazon\_reviews\_multi}         & 69.17\% & 24.80\% & 31.80\%    & 34.39\%      & 22.84\%       & 31.80\%    & 55.94\%$^{\dagger}$   \\
\texttt{poem\_sentiment}                & 74.31\% & 26.34\% & 58.03\%    & 45.32\%      & 22.84\%       & 47.54\%    & 73.56\%      \\
\texttt{tweet\_eval-emoji}              & 74.31\% & 23.26\% & 63.28\%    & 23.45\%      & 44.88\%       & 79.02\%    & 55.94\%      \\
\texttt{tweet\_eval-hate}               & 61.46\% & 23.26\% & 26.55\%    & 61.73\%      & 44.88\%       & 47.54\%    & 55.94\%      \\
\texttt{tweet\_eval-irony}              & 61.46\% & 23.26\% & 26.55\%    & 23.45\%      & 22.84\%       & 26.55\%    & 64.75\%       \\
\texttt{tweet\_eval-offensive}          & 61.46\% & 23.26\% & 26.55\%$^{\dagger}$ & 34.39\%      & 28.35\%       & 31.80\%    & 55.94\%       \\
\texttt{tweet\_eval-femminist}          & 61.46\% & 23.26\% & 26.55\%    & 39.05\%      & 22.84\%       & 37.05\%    & 64.75\%       \\
\hline
\end{tabular}
\end{adjustbox}
\caption{Results obtained from the analysis of additional datasets not shown in Tab.\ref{tab:Results}. The values presented in this table correspond to the lowest percentage of reset parameters that KEN achieved without impacting the model performance. The ${\dagger}$ symbol denotes a reset parameter rate that falls below the minimum value reported in Tab. \ref{tab:Results}}
\label{tab:Additional_results}
\end{table*}
\newpage
\section{Additional results} \label{Apx: Additional_results}
This appendix presents additional results obtained using KEN that are not included in Tab. \ref{tab:Results}.

Tab. \ref{tab:All datasets} provides a comprehensive overview of all datasets analyzed in the paper. In contrast to Tab. \ref{tab:Results}, Tab. \ref{tab:Additional_results} focuses on the specific results included in this appendix. Here, we highlight cases where KEN achieves F1-weighted scores that match or surpass the original unpruned model performance, focusing on the highest percentage of reset parameters for each model on each dataset. This emphasizes KEN remarkable compression capabilities, allowing it to achieve comparable or even improved performance while significantly reducing the model parameter count.
\begin{table}[htb!]
    \centering
    \begin{adjustbox}{width=\linewidth}
    \begin{tabular}{lr}
    \hline
    Dataset  &  Reference \\ \hline
    \texttt{trec} & \citealp{li-roth-2002-learning} \\
    \texttt{AG-NEWS} & \citealp{gulliAgnews} \\
    \texttt{rotten tomatoes} & \citealp{pang2005seeing} \\
    \texttt{IMDB} & \citealp{maas-etal-2011-learning} \\
    \texttt{ade\_corpus\_v2} & \citealp{ade} \\
    \texttt{glue-sst2} & \citealp{socher-etal-2013-recursive} \\
    \texttt{YELP POLARITY} & \citealp{zhang2015character} \\
    \texttt{hate\_speech\_offensive} & \citealp{davidson2017automated} \\
    \texttt{hate\_speech18} & \citealp{gibert2018hate} \\
    \texttt{EMO} & \citealp{chatterjee-etal-2019-semeval} \\
    \texttt{scicite} & \citealp{cohan-etal-2019-structural} \\
    \texttt{amazon\_reviews\_multi} & \citealp{marc_reviews} \\
    \texttt{poem sentiment} & \citealp{sheng-uthus-2020-investigating} \\
    \texttt{tweet\_eval-emoji} & \citealp{barbieri-etal-2020-tweeteval} \\
    \texttt{tweet\_eval-hate} & \citealp{barbieri-etal-2020-tweeteval} \\
    \texttt{tweet\_eval-irony} & \citealp{barbieri-etal-2020-tweeteval} \\
    \texttt{tweet\_eval-offensive} & \citealp{barbieri-etal-2020-tweeteval} \\
    \texttt{tweet\_eval-feminist} & \citealp{barbieri-etal-2020-tweeteval} \\
    
    \end{tabular}
    \end{adjustbox}
    \caption{Dataset analyized}
    \label{tab:All datasets}
\end{table}
\newpage
\section{KEN$_{viz}$ examples} \label{Apx: KEN_viz}
KEN$_{viz}$ generates visual representations of the model pruning after the KEN application. Here, we focus on key matrices from layers 0 and 12 of a BERT model trained on the \texttt{glue-sst2} dataset (details in Sec. \ref{subsec: result and comparison}). For each layer, we present both a single matrix view and a neighbor count view, as described in Sec. \ref{subsec:KEN_viz}.

BERT was chosen for this experiment due to its exceptional performance across a range of k values during testing (Tab. \ref{tab:Results} and Tab. \ref{tab:Additional_results}). To comprehensively explore how parameter selection patterns evolve, we employed three different $k$ values, representing varying degrees of parameter selection. This allowed us to observe how parameter choices shift as the amount of parameter resetting increases.

\begin{figure}[t!]
    \centering
    \begin{subfigure}[b]{\linewidth}
        \centering
        \includegraphics[width=0.475\linewidth]{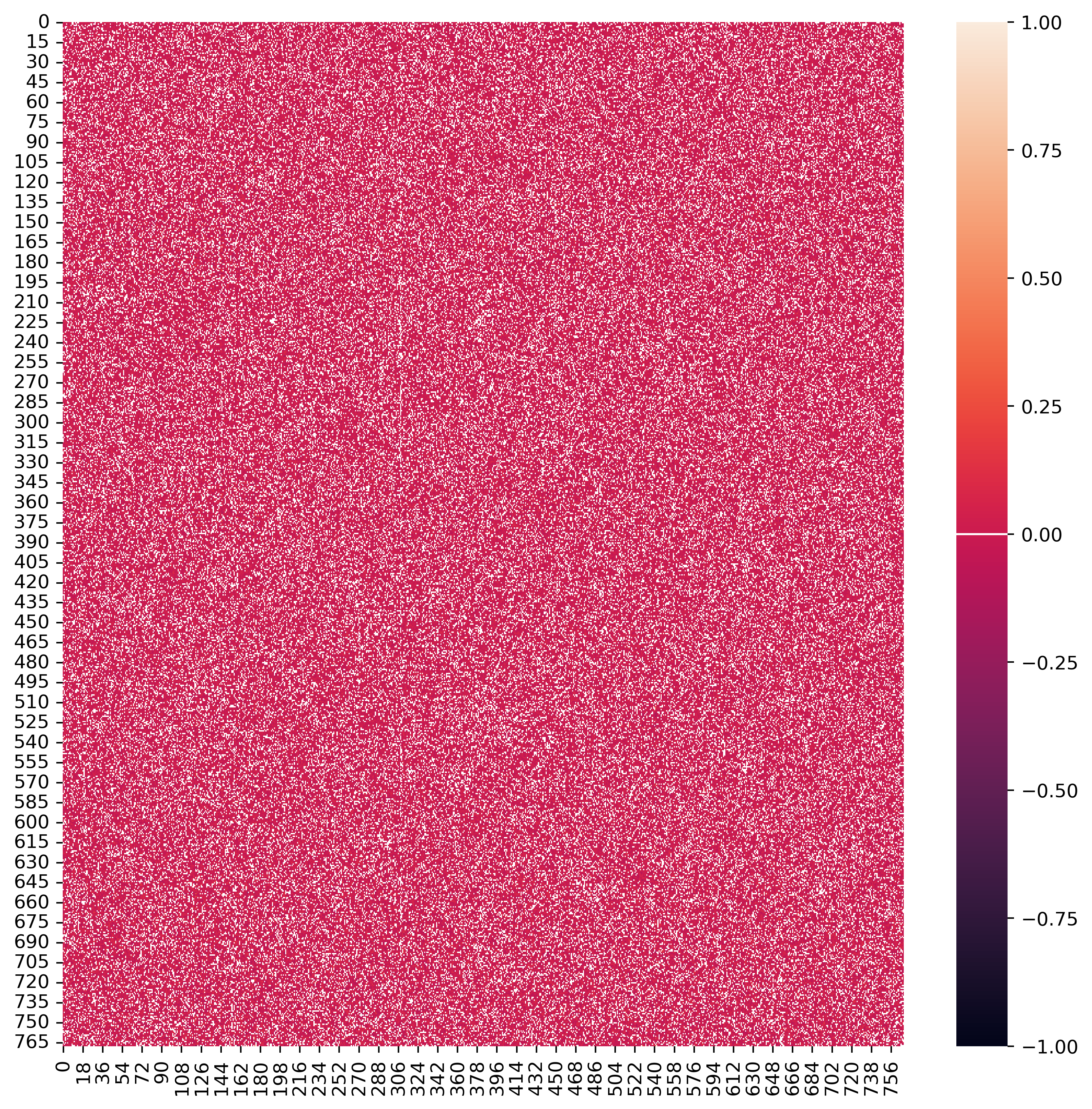}%
        \hfill
        \includegraphics[width=0.475\linewidth]{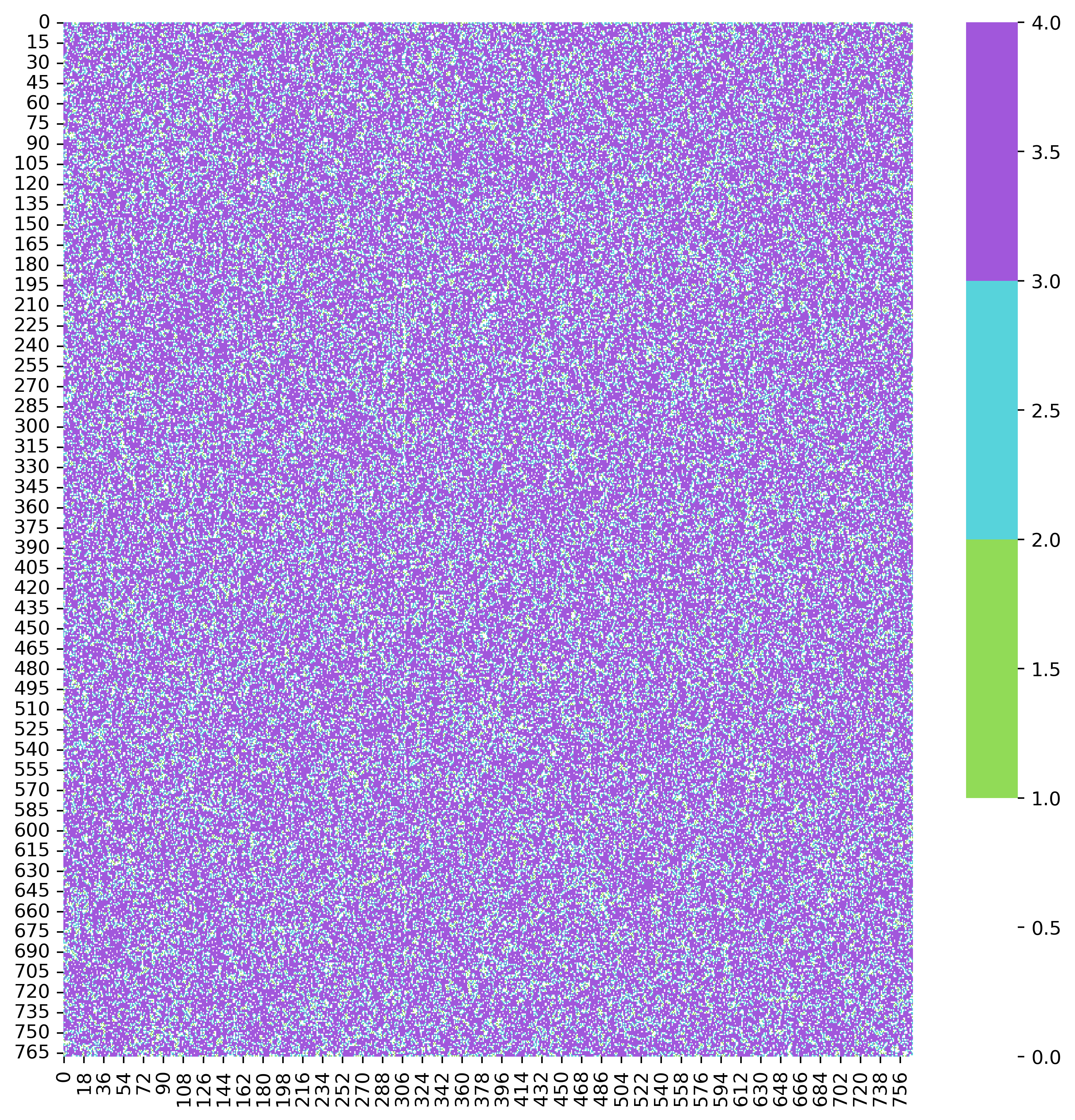}
        \caption{Parameter reset 21.87\%}
    \end{subfigure}
    \vskip\baselineskip
    \begin{subfigure}[b]{\linewidth}
        \centering
        \includegraphics[width=0.475\linewidth]{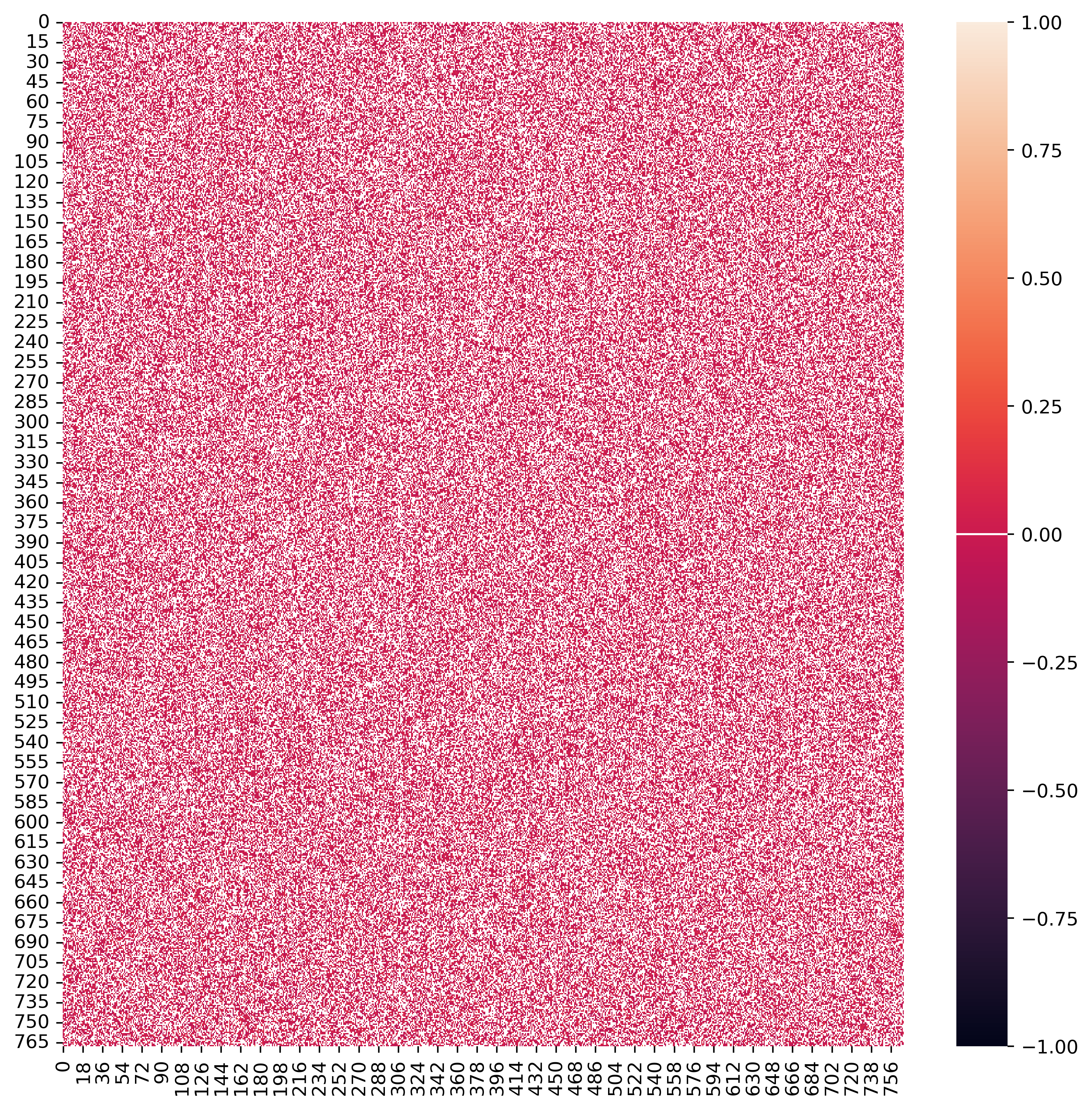}%
        \hfill
        \includegraphics[width=0.475\linewidth]{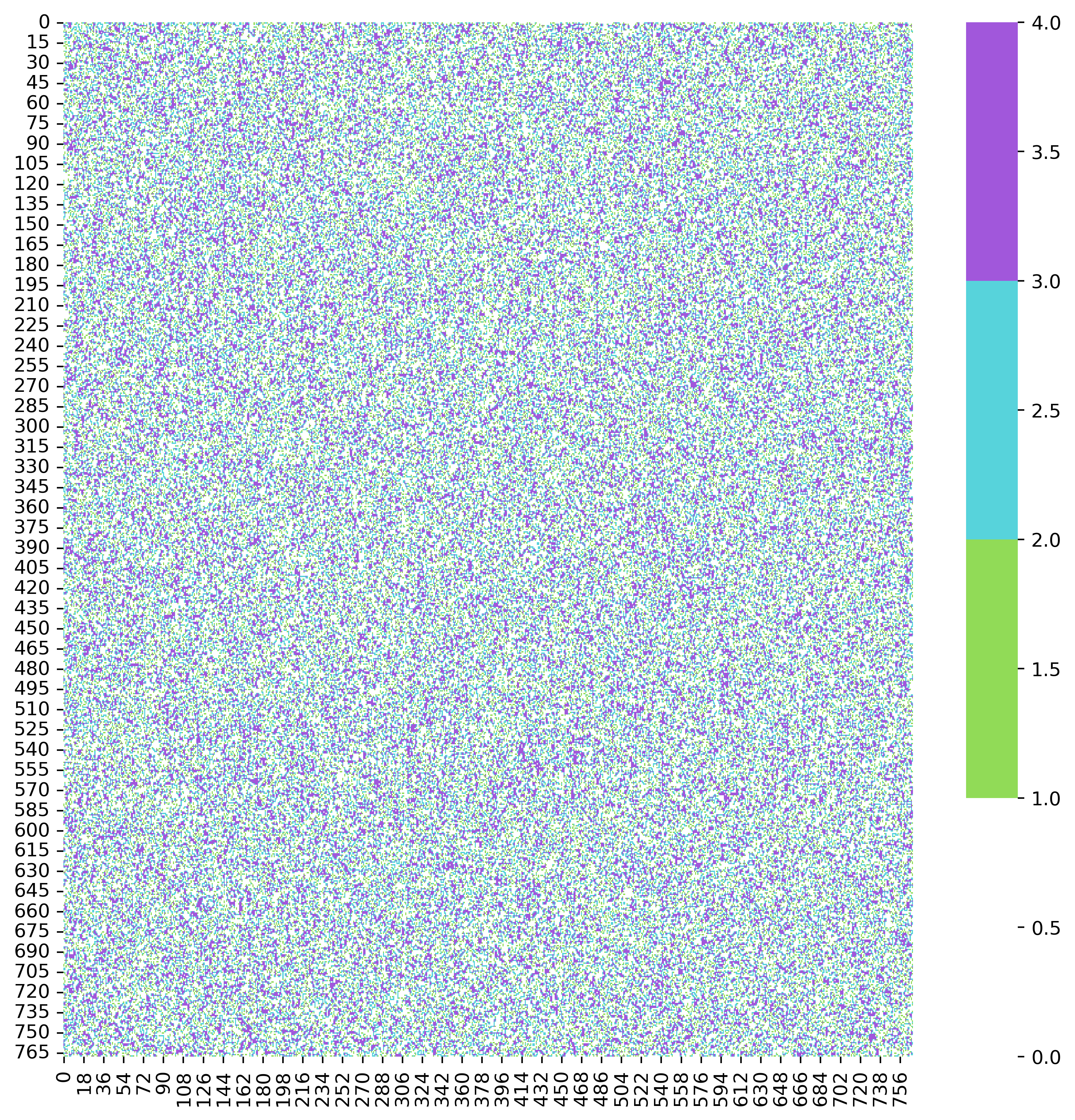}
        \caption{Parameter reset 47.91\%}
    \end{subfigure}

    \vskip\baselineskip
    \begin{subfigure}[b]{\linewidth}
        \centering
        \includegraphics[width=0.475\linewidth]{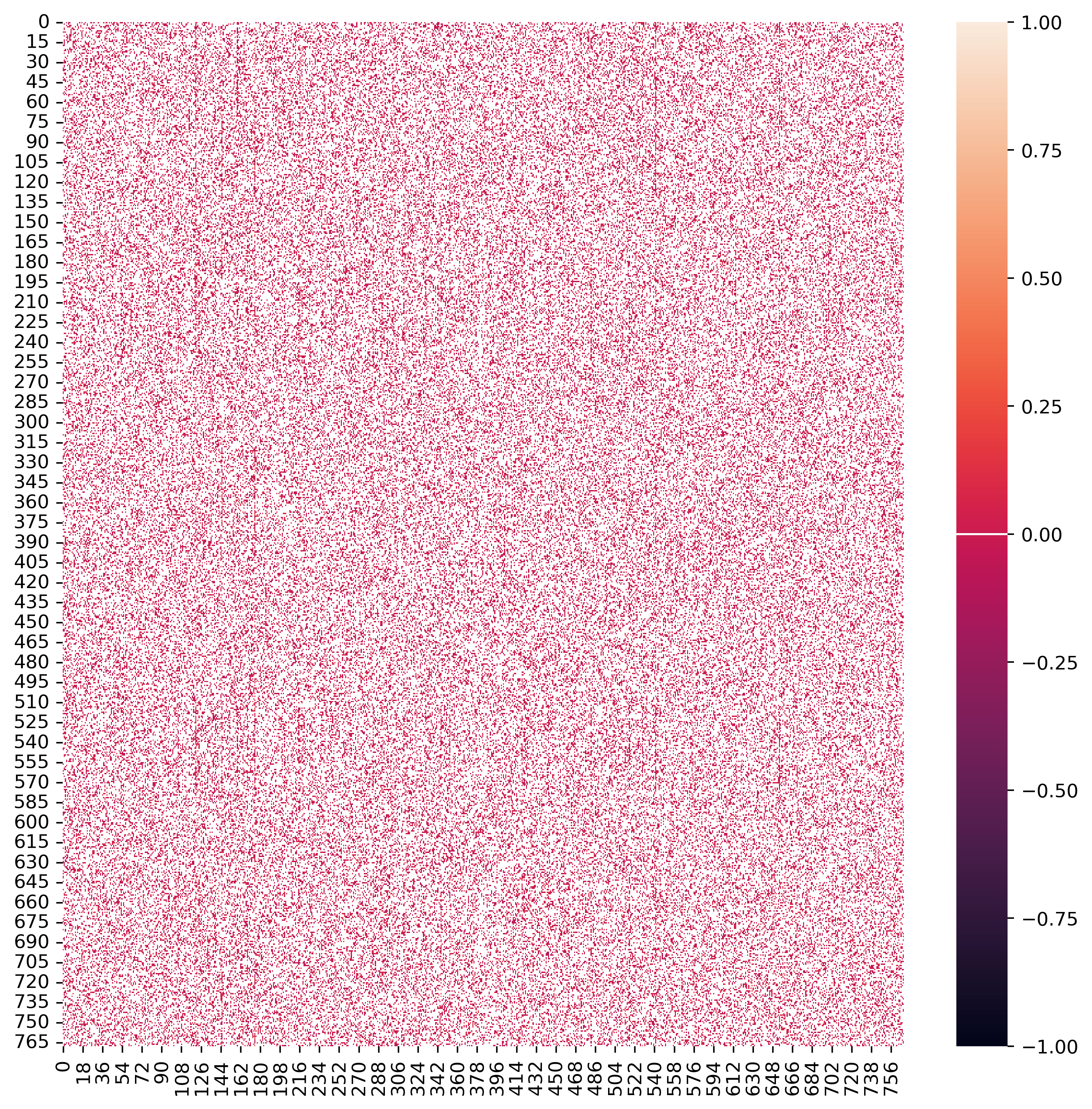}%
        \hfill
        \includegraphics[width=0.475\linewidth]{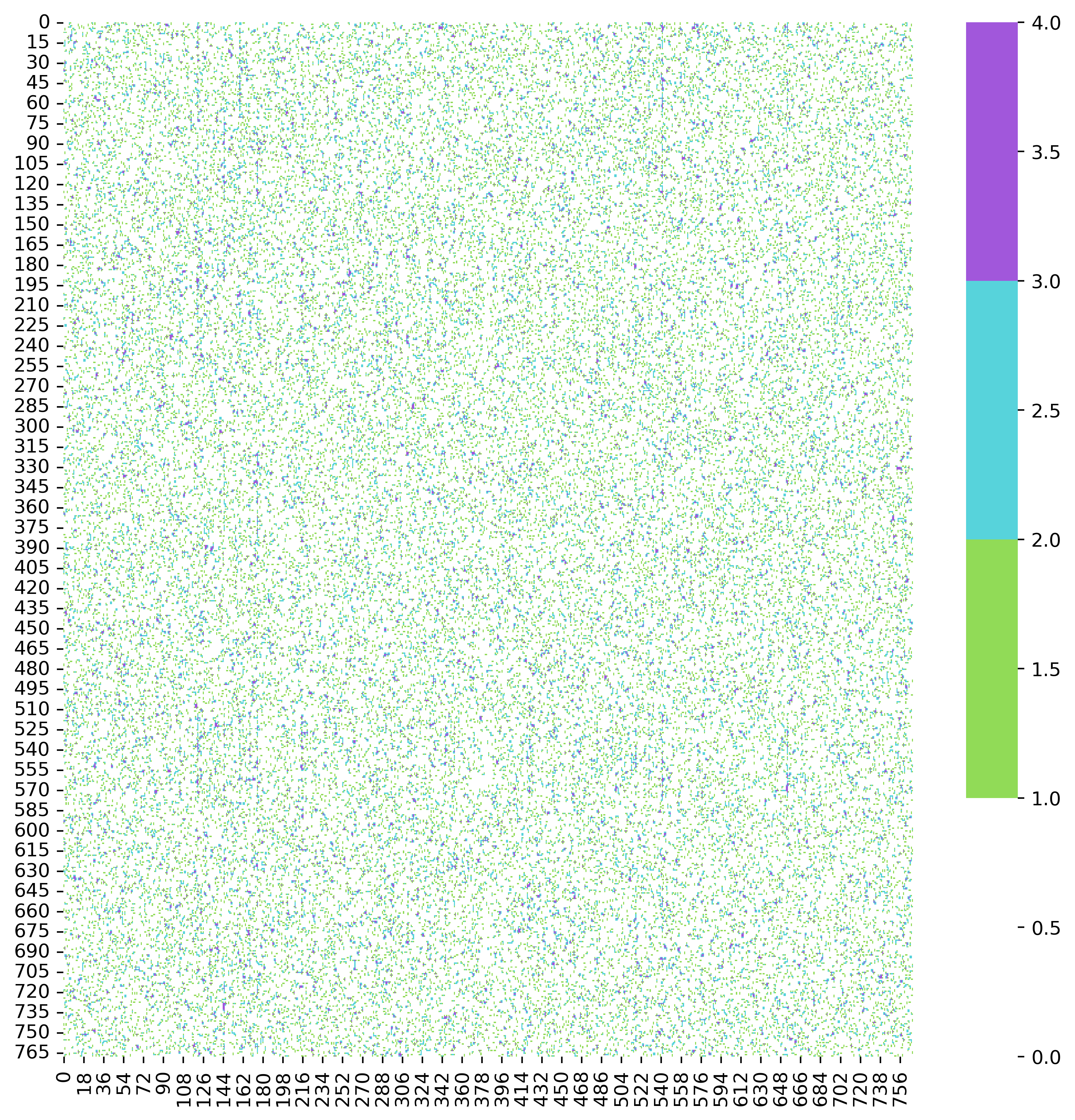}
        \caption{Parameter reset 73.95\%}
    \end{subfigure}
    \caption{KEN$_{viz}$ visualization of the key attention matrix at \underline{layer 0} of a BERT model trained on the \texttt{glue-sst2} dataset. The left-hand figures depict the matrix after undergoing the KEN pruning stage, while the right-hand ones showcase the corresponding neighbor counts}
    \label{fig: key layer 0}
\end{figure}

\begin{figure}[t!]
    \centering
    \begin{subfigure}[b]{\linewidth}
        \centering
        \includegraphics[width=0.475\linewidth]{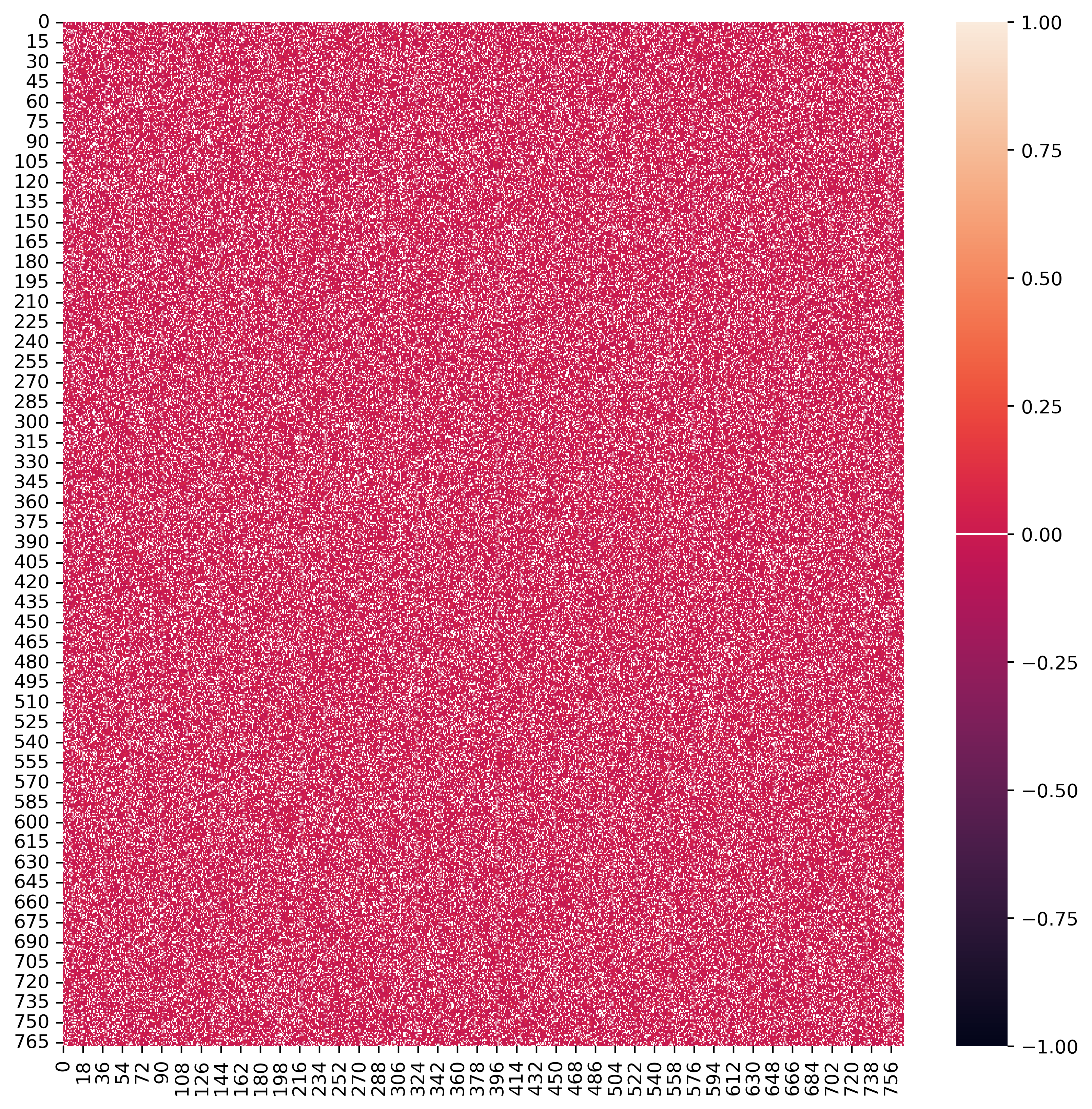}%
        \hfill
        \includegraphics[width=0.475\linewidth]{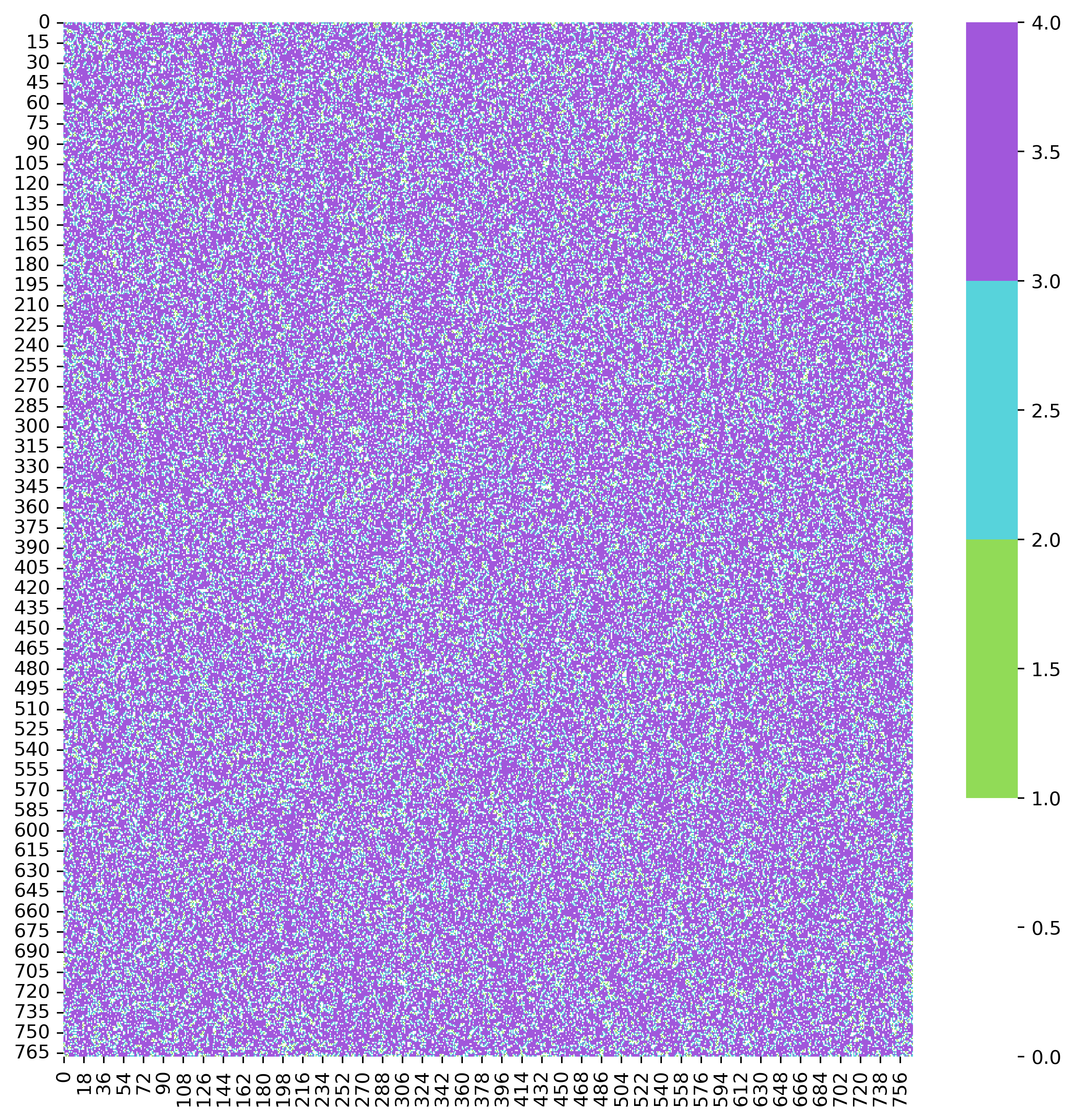}
        \caption{Parameter reset 21.87\%}
    \end{subfigure}
    \vskip\baselineskip
    \begin{subfigure}[b]{\linewidth}
        \centering
        \includegraphics[width=0.475\linewidth]{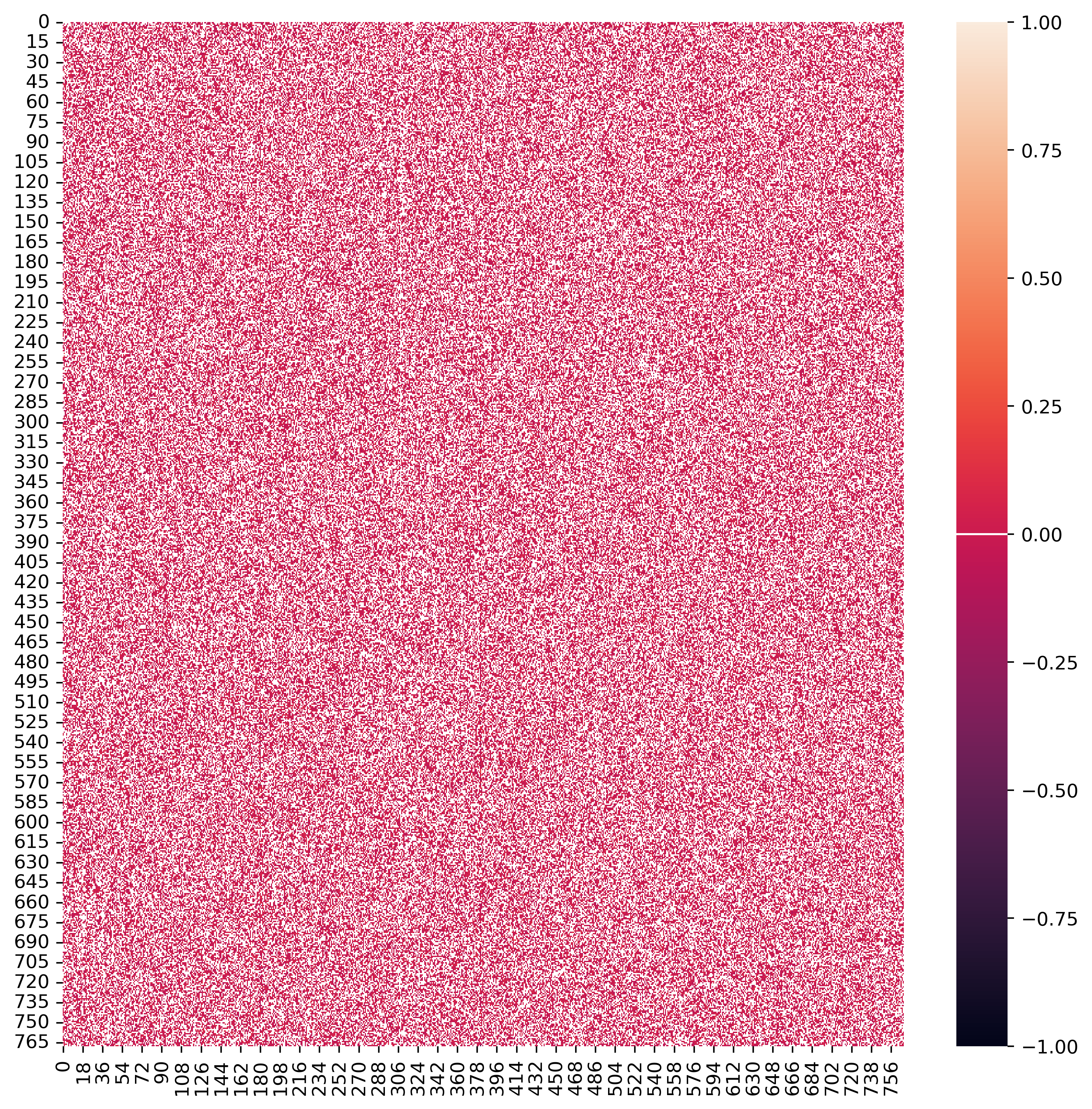}%
        \hfill
        \includegraphics[width=0.475\linewidth]{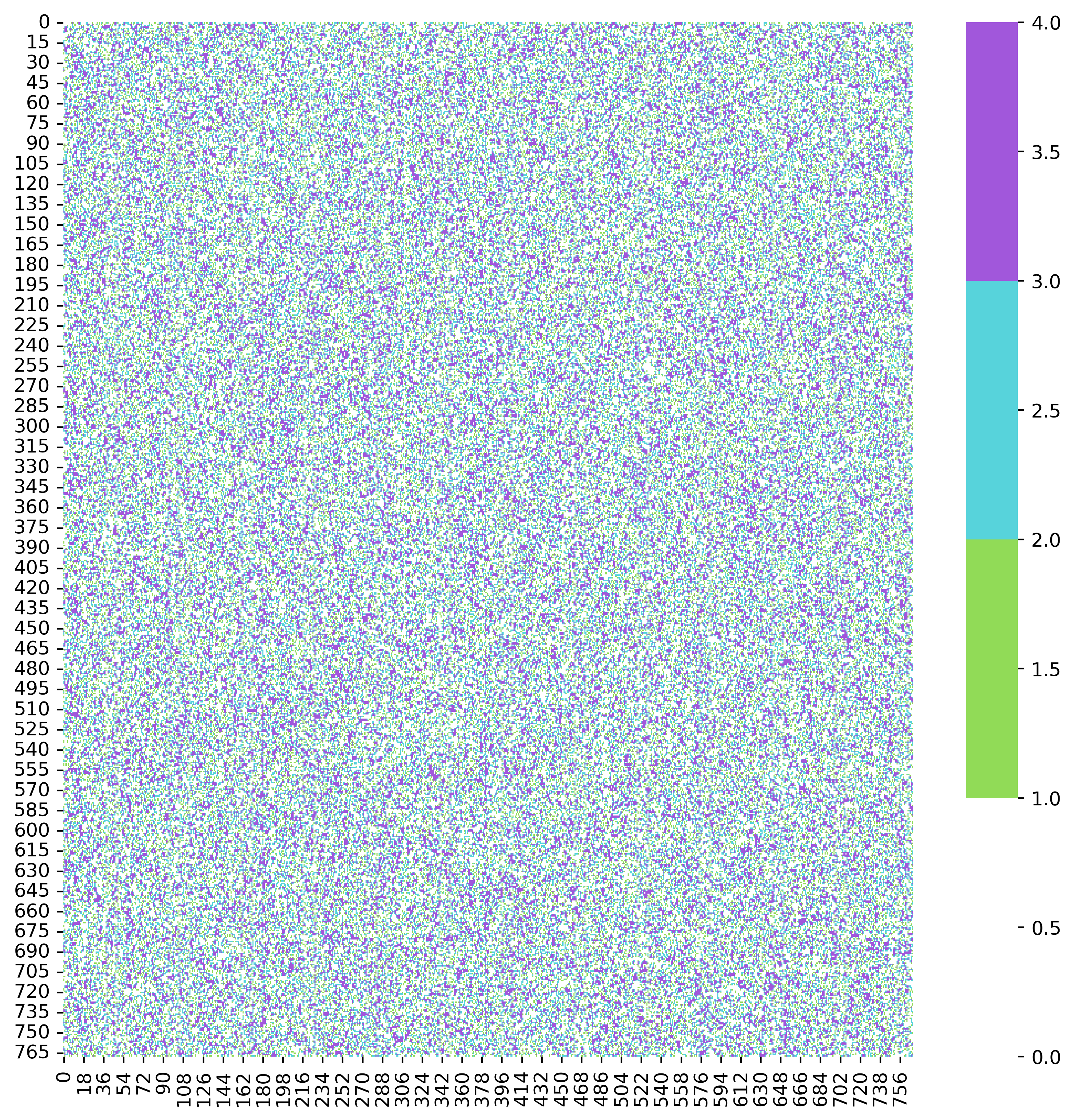}
        \caption{Parameter reset 47.91\%}
    \end{subfigure}

    \vskip\baselineskip
    \begin{subfigure}[b]{\linewidth}
        \centering
        \includegraphics[width=0.475\linewidth]{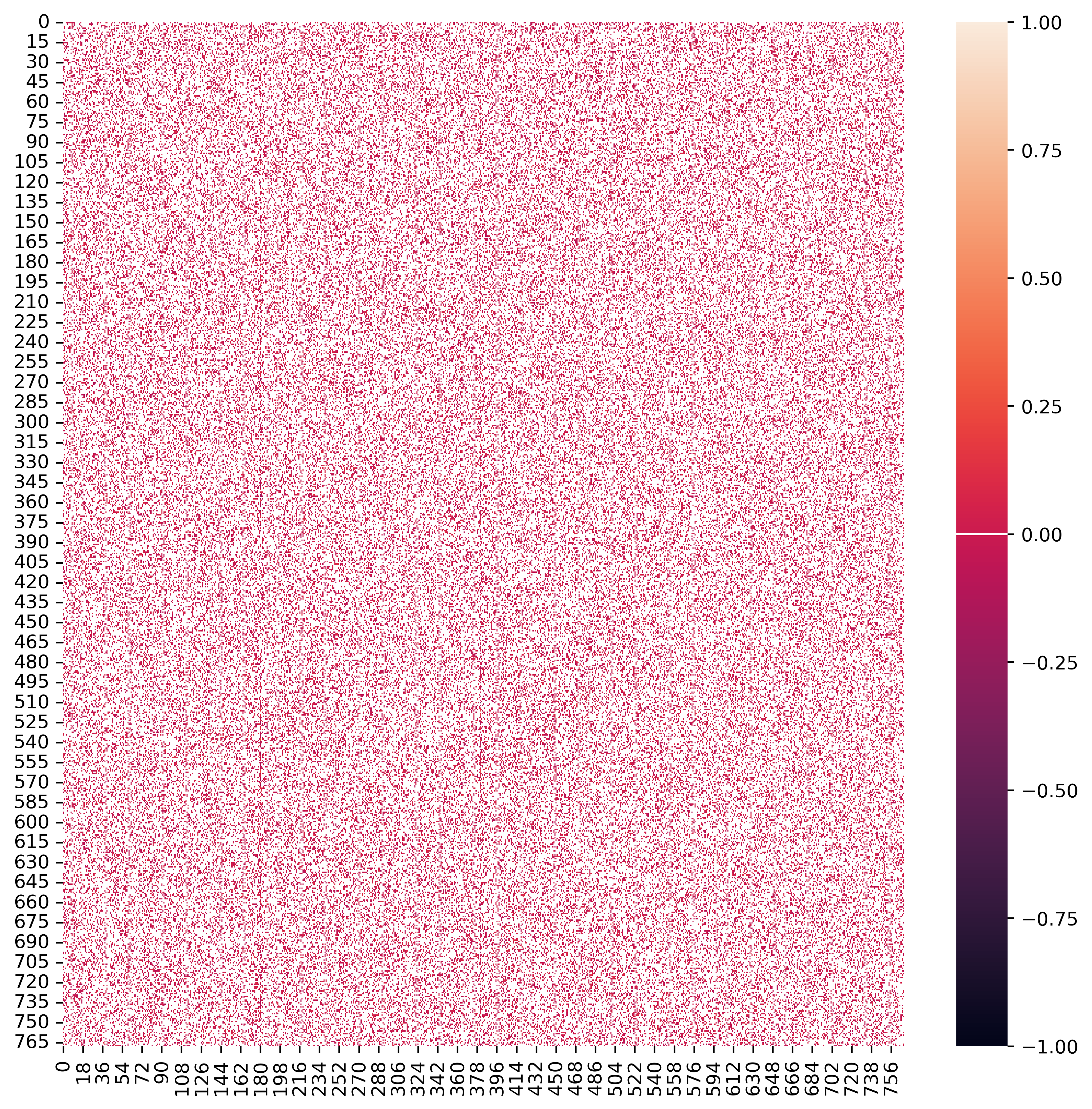}%
        \hfill
        \includegraphics[width=0.475\linewidth]{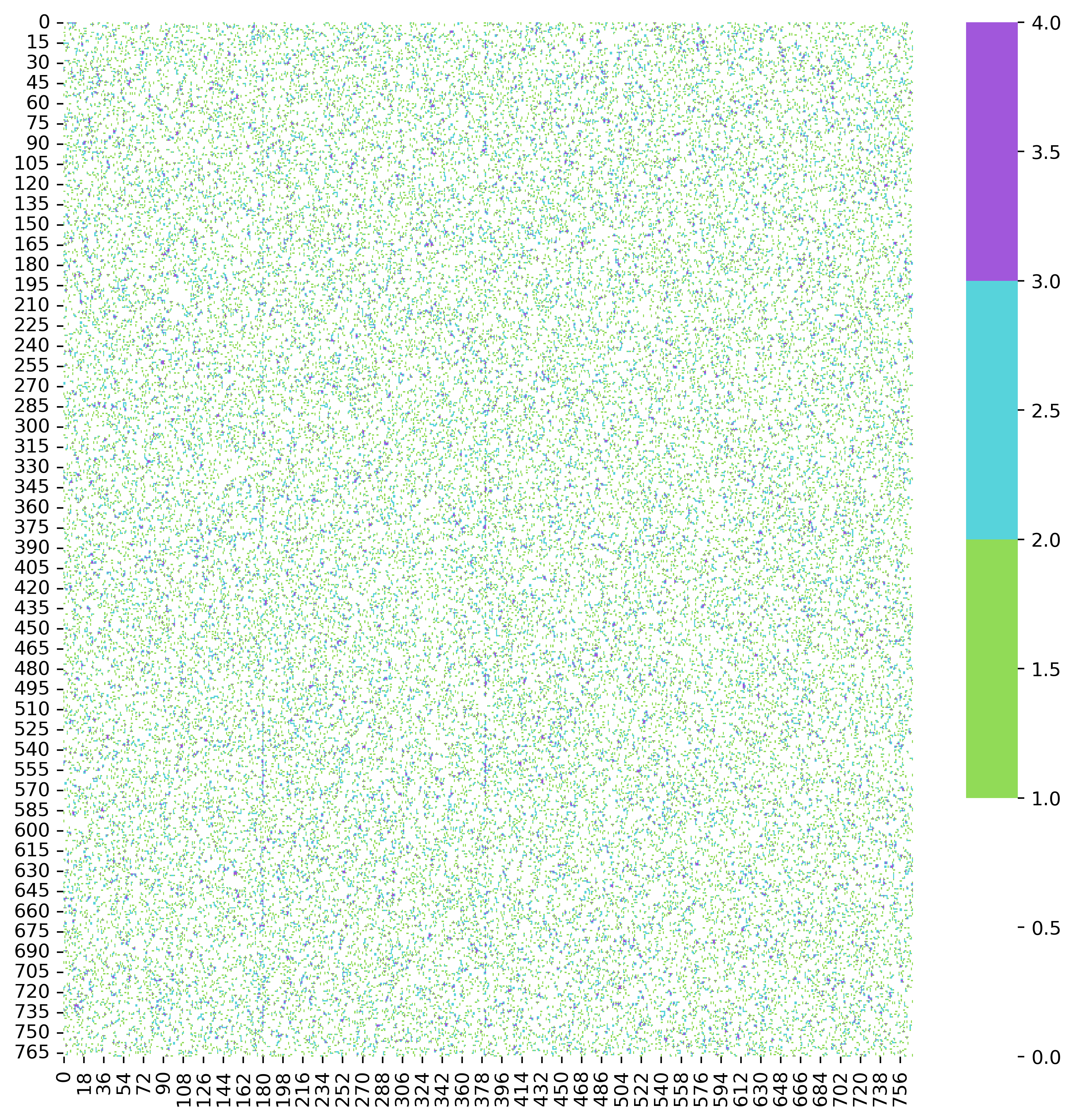}
        \caption{Parameter reset 73.95\%}
    \end{subfigure}
    \caption{KEN$_{viz}$ visualization of the key attention matrix at \underline{layer 12} of a BERT model trained on the \texttt{glue-sst2} dataset. The left-hand figures depict the matrix after undergoing the KEN pruning stage, while the right-hand ones showcase the corresponding neighbor counts}
    \label{fig: key layer 12}
\end{figure}

Fig. \ref{fig: key layer 0} and Fig. \ref{fig: key layer 12} consistently reveal a uniform distribution of parameters within each matrix row across all configurations and layers. This implies an absence of well-defined clusters of selected parameters. Furthermore, the number of neighbors for each parameter remains consistent regardless of the chosen $k$ value.

\end{document}